\documentclass[letterpaper, 10 pt, conference]{ieeeconf}
\IEEEoverridecommandlockouts
\usepackage[utf8]{inputenc}
\usepackage{pgfplots}
\DeclareUnicodeCharacter{2212}{−}
\usepgfplotslibrary{groupplots,dateplot}
\usetikzlibrary{patterns,shapes.arrows}
\pgfplotsset{compat=newest}

\usepackage{xspace}
\usepackage{amsmath}
\usepackage{mathtools}
\usepackage{nicefrac}
\usepackage{amsfonts}
\usepackage{pifont}
\usepackage{floatflt}
\usepackage{url}
\usepackage{dsfont}

\usepackage{tikz}
\usetikzlibrary{calc, positioning, shadows, backgrounds, arrows.meta}
\usepackage{multirow}
\usepackage{booktabs}
\usepackage{tabularx}
\newcolumntype{L}{>{\raggedright\arraybackslash}X}

\usepackage{algorithm}

\makeatletter
\DeclareRobustCommand*\cal{\@fontswitch\relax\mathcal}
\makeatother

\newcommand{\commentout}[1]{}

\newcommand{\V}{{\cal V}}

\newcommand{\sat}{\models}

\newcommand{\lem}{\begin{lemma}}
\newcommand{\elem}{\end{lemma}}
\newcommand{\pro}{\begin{proposition}}
\newcommand{\epro}{\end{proposition}}

\newcommand{\dfn}{\begin{definition}}
\newcommand{\edfn}{\end{definition}}

\newcommand{\xam}{\begin{example}}
\newcommand{\exam}{\end{example}}

\usepackage{algorithm}

\usepackage{longtable}
\usepackage{subcaption}
\usepackage{pgfplots}
\usepackage{pgfplotstable}
\usepackage{paralist}
\usepgfplotslibrary{fillbetween}
\pgfplotsset{compat=1.16}

\newcommand{\?}{\stackrel{?}{=}}

\usepackage{comment}
\usepackage{graphicx}
\usepackage{hyperref}
\usepackage{xcolor}
\let\labelindent\relax
\usepackage{enumitem}
\usepackage[noend]{algpseudocode}
\algrenewcommand\algorithmicindent{0.6em}%
\usepackage[capitalize,noabbrev]{cleveref}
\usepackage{pgfplots}
\pgfplotsset{compat=1.18}
\usepackage{tikz}
\usetikzlibrary{fit}
\usepackage{xspace}
\usepackage{orcidlink}

\crefname{line}{line}{lines}
\crefname{figure}{Fig.}{Figs.}
\Crefname{figure}{Fig.}{Figs.}
\crefname{equation}{Eq.}{Eqs.}
\Crefname{equation}{Eq.}{Eqs.}
\crefname{section}{Sec.}{Secs.}
\Crefname{section}{Sec.}{Secs.}
\crefname{definition}{Def.}{Defs.}
\Crefname{definition}{Def.}{Defs.}
\crefname{algorithm}{Alg.}{Algs.}
\Crefname{algorithm}{Alg.}{Algs.}
\Crefname{algocf}{Alg.}{Algs.}
\Crefname{appendix}{Appendix}{Appendices}

\newtheorem{example}{Example}
\newtheorem{lemma}{Lemma}

\newtheorem{definition}{Definition}

\newtheorem{proposition}{Proposition}

\newcommand{\status}{\mathtt{status}}

\newcommand{\state}{\xi}
\newcommand{\statetraj}{\overline{\state}}

\newcommand{\env}{\textit{env}}

\newcommand{\aref}[1]{\hyperref[#1]{Appendix~\ref*{#1}}}

\newcommand{\KEnote}[1]{}

\DeclareMathOperator{\nec}{nec}
\DeclareMathOperator{\suf}{suf}

\usepackage{comment}

\begin{document}
\graphicspath{{./images/}}


\title{\LARGE \bf
Sufficient and Necessary Causal Explanations\\for the Analysis of Autonomous Systems 
}

\author{Authors omitted for anonymous review}

\author{
Tom Yaacov$^{*,1}$\orcidlink{0000-0002-0565-6506}, Khen Elimelech$^{1}$\orcidlink{0000-0003-1391-8956}, Hana Chockler$^{1}$\orcidlink{0000-0003-1219-0713}, and Moshe Y. Vardi$^{2}$\orcidlink{0000-0002-0661-5773} 
\thanks{$^{1}$ King's College London,
        {\tt\footnotesize \{firstname.lastname\}@kcl.ac.uk}}%
\thanks{$^{2}$ Rice University,
        {\tt\footnotesize vardi@rice.edu}}%
\thanks{* Corresponding Author.}
\thanks{** This work was partially supported by the Causality in Healthcare AI (CHAI) Hub [UKRI AI and EPSRC grant EP/Y028856/1].}
}

\maketitle

\begin{abstract}
Understanding failures of autonomous systems is challenging, particularly when their behavior depends on black-box components such as neural networks. In this work, we make a distinction between two objectives of failure analysis: reproduction, which seeks to identify conditions sufficient to reproduce an observed failure, and avoidance, which seeks to identify the changes required to prevent it. Previous work on causal explanations of system failures is based on the concept of sufficient explanations and naturally supports reproduction. Nevertheless, we observe that it does not address the problem of avoiding the failure. In this paper, we introduce and formalize the concept of necessary explanations for system analysis as the correct formalism for failure avoidance. We place necessary explanations in the framework of actual causality and show that, together with sufficient explanations, they provide a unified causal framework for addressing both objectives of failure analysis. We present an efficient approximation algorithm for generating necessary explanations based on the notion of causal responsibility. We evaluate the efficiency of our approach across multiple environments. We demonstrate the practical utility of necessary explanations for failure avoidance through counterfactual generation and show that it outperforms existing methods. 


\end{abstract}

\section{Introduction}
Autonomous systems rely on algorithms and digital computation to control real-world machines and achieve complex objectives autonomously~\cite{cps}. As these objectives become increasingly challenging, modern systems have come to rely heavily on black-box components, such as neural networks~\cite{varshney2019deepcontrol,wang2021bounding}. The opacity of these components, combined with the complexity of systems and the variability of their operating environments, makes it difficult to interpret system behavior and gain insight into how decisions were made.
This difficulty has motivated the rapidly growing interest in Explainable Artificial Intelligence (XAI), which is widely regarded as crucial for the widespread adoption of autonomous systems~\cite{hoenig2024explainable,moosavi2024explainable}. 

Explainability becomes even more vital when failures happen. Understanding failures, known as \emph{failure analysis}, is crucial for the development and debugging of systems, and it is especially challenging when systems rely on black-box components, often provided by a third party. 
An explanation should help us gain insight into what affected the system behavior that led to the failure outcome. Addressing this goal requires answering two questions: (1) what counts as an explanation? (2) how to generate the right one? 
Post-hoc XAI, where explanations are based on the analysis of the outcome rather than the process, 
is a suitable paradigm for analyzing black-box systems, as it does not require any changes or even access to 
the internals of the system, and it is the one we follow in this paper. 

To position this work clearly, we first mention the underlying conceptual contribution of this paper (which we discuss in detail in \cref{sec:concept}), with which we identify the two complementary explanation objectives: \emph{outcome avoidance}, and \emph{outcome reproduction}. In the first case, our concern is \emph{changing} the outcome (with minimal effort), e.g., when we wish to \emph{repair} the failure; the explanation should hence provide insight on which elements in the scenario should be changed (and how), in order to change the (failing) outcome.
In the second case, our concern is generating additional scenarios, in which the same (failing) outcome is \emph{maintained}, e.g., as part of a debugging process; the explanation should hence provide insight on which elements in the scenario are sufficient to recreate the outcome, regardless of the rest. The two forms of explanation are complementary, but one cannot be deduced from the other.


A recent work~\cite{icra} suggested using the framework of ``actual causality''~\cite{Hal16,CH24} 
to formally define and generate \emph{causal} failure explanations (for CPSs). 
According to our categorization, that work addressed the objective of failure reproduction, by focusing on the generation 
of ``sufficient explanations'' (an established concept in actual causality). Roughly speaking, a sufficient explanation is a subset of elements
of the failure execution that is, by itself, sufficient to result in a failure, even if all other elements are changed or removed.
Such explanations are not suitable for failure avoidance, as removing or changing such a sufficient subset does not guarantee that the failure
is avoided. Intuitively, this is because multiple sufficient subsets may exist.

This work seeks to extend the foundations laid in~\cite{icra} to address the objective of failure avoidance by introducing and applying the concept of ``necessary explanations''. Roughly speaking, a necessary explanation is a subset of elements of the failure execution such that
removing them avoids the failure. Together, this provides, for the first time, a unified (causal) formal framework for failure analysis, allowing to address both objectives. As exact computation of such causal explanations quickly becomes intractable when the number of inputs increases~\cite{CKKS26}, we also develop a practical, scalable algorithm
for generating necessary explanations based on the notion of causal \emph{degree of responsibility}~\cite{CH04}.

\begin{figure}[t]
\centering
\includegraphics[width=\columnwidth]{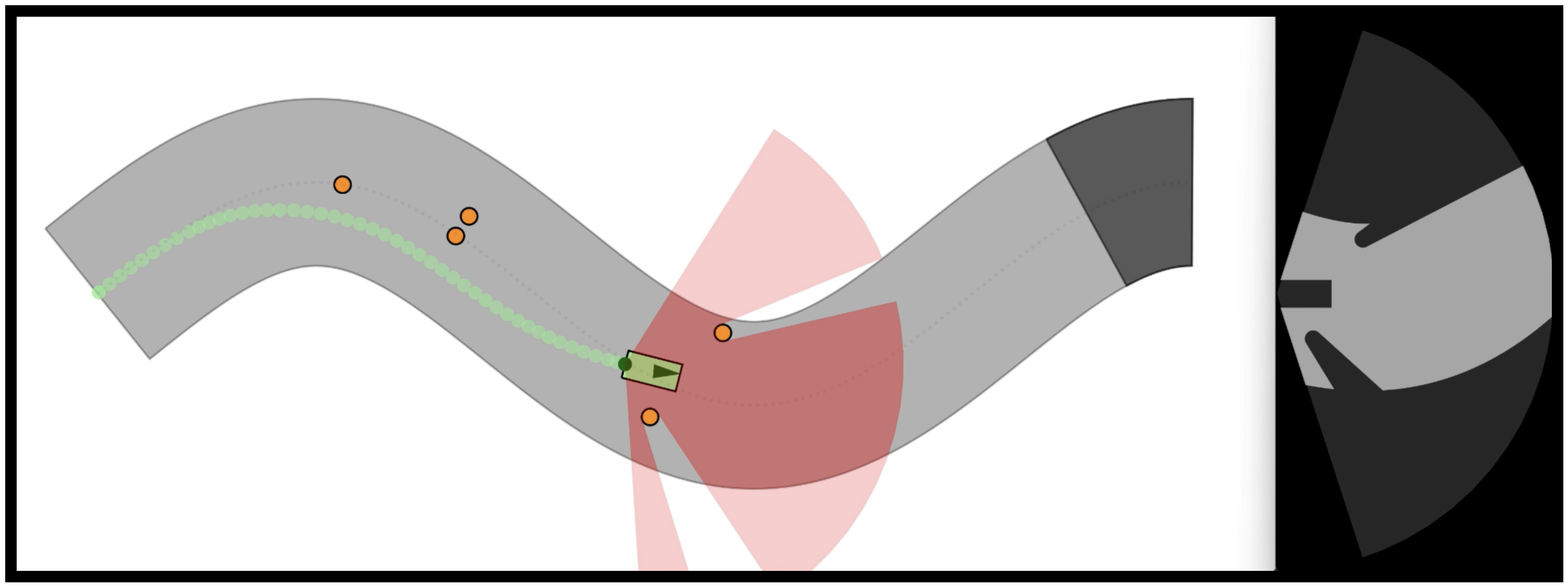}
\caption{The autonomous car in the LiteRacer simulator~\cite{Elimelech24LiteRacer}. 
}
\label{fig:car}
\end{figure}

\textbf{The paper contributions:}  
\textbf{(i)} We suggest a conceptual categorization of the objectives of failure explanation---avoidance and reproduction---providing a basis for understanding, 
comparing, and advancing work in this field.
\textbf{(ii)} We introduce the notion of necessary explanations for autonomous system failures and discuss its application to failure avoidance. 
\textbf{(iii)} We develop a practical, scalable algorithm for generating necessary explanations and provide an experimental evaluation demonstrating its efficiency across multiple environments.
Finally, to demonstrate the superiority of this framework for explanation for failure avoidance, \textbf{(iv)} we provide a study comparing 
with prior work on failure avoidance not based on causal reasoning. We show that our approach achieves a higher success rate, lower computation time, and better quality.
For clarity, the paper is presented (and contributions (iii) and (iv) are evaluated) using a setup of an autonomous car with a black-box neural controller, trained to avoid collisions (as in \cref{fig:car}). In our evaluation, we use two autonomous driving simulators, of both high and low fidelity, showing that the benefits of our approach are simulator- and system-independent.

The rest of the paper is structured as follows. 
After overviewing the related work,
\cref{sec:prelim} provides preliminaries on the autonomous system testing model and introduces the running example.
\cref{sec:fundamentals} gives a brief overview of actual causality and introduces our setup.
\cref{sec:concept} introduces the conceptual framework of failure analysis.
\cref{sec:ac-for-cps} presents our theoretical contributions, extending the framework with necessary explanations. \cref{sec:algorithm} presents the approximation algorithm for computing necessary explanations. The experimental performance evaluation is presented in \cref{sec:performance-evaluation}. \cref{sec:counterfactuals-evaluation} demonstrates the practical utility of necessary explanations for failure avoidance, comparing it with existing work. \cref{sec:conclusion} concludes the paper.

\noindent\textbf{Related Work:} 
In explainable AI, actual causality has been used in various domains, such as images~\cite{CKKS26}, audio~\cite{kelly_i_2026}, and process mining~\cite{yaacov2026causal}. Since computing actual causes is generally hard, these approaches introduce efficient approximations~\cite{CKK25}. Similarly to us, they treat the AI model as a black box.

Existing explainability approaches for failure reproduction include sensitivity analysis~\cite{narteni2022sensitivity,diwakaran2017}, which examines neighborhoods around falsifying traces, identifying regions where a property is violated. As mentioned, actual causality has also been proposed for failure reproduction, providing a more structured, high-level alternative by identifying minimal subsets of elements that reproduce the observed failure~\cite{icra}.

Failure avoidance methods focus on modifications to a failing scenario that change its outcome. Existing approaches for failure avoidance primarily generate \emph{counterfactuals}, passing examples similar to the original~\cite{ghazal2026towards,manca2023explainable}. 
While counterfactuals can provide valuable insight into the observed behavior, they focus on input-output changes rather than the underlying causal mechanisms, potentially attributing importance to downstream effects and including modifications that are not necessary for avoiding the failure. 
Autonomous systems work on explainability also use counterfactual attribution for root-cause analysis~\cite{fang2026counterfactual}. This approach relies on causal discovery, which infers causal relationships from data,
a fundamentally different approach from
the one we adopt in this paper, which allows reasoning within an abstract model. 

\section{Preliminaries}
\label{sec:prelim}


To facilitate analysis of autonomous systems, we adopt the formalism introduced in~\cite{elimelech2024falsification} and later extended in~\cite{icra}. For generality, this formalism deals with systems with time-extended operation, whose output is a trajectory (like Cyber-Physical Systems); though, one can consider a trajectory consisting of a single, final state.

An environment~$\env$ defines the set of variables needed to describe the surroundings of a system, capturing all the external factors affecting its operation. 
The system operation is determined by observing and potentially interacting with the (static) environment.
We use~$\state_t$ to denote the system state at time~$t$, and~$\statetraj$ to denote the concatenation of states from start to finish, a time-parameterized trajectory produced by the system execution in~$\env$, starting from an initial system state~$\state_0$.
Without loss of generality, to simplify the discussion, we may consider the initial state to be predetermined; this means we consider the environment to be the only source of trajectory variability, with no other uncertainty sources. 

We do not require access to an explicit model of the system (i.e., it is a ``black-box''), but assume it can be simulated; that is, given an environment specification (and an initial state), we can simulate the system to generate a trajectory.
We assume a rudimentary understanding of the system, allowing us to differentiate between the environment \emph{elements}, which are the ``optional'' variables used to enrich the environment, and environment \emph{parameters}, which are mandatory variables.

We test and analyze the system in the context of a \emph{task}, which the system was designed to complete, in relation to its environment. 
Accordingly, we define a testing model that wraps the (black-box) system; its input is an environment $\env$, and its output is an indicator of task success, based on the system trajectory (the simulation output).
For correct analysis, we care not only to differentiate between success and failure in the task, but also between different types of failures. To evaluate this, we assume the availability of a ``status'' predicate:
\begin{equation}
    \status(\statetraj,\env) \in \{1, -1, -2, -3, \dots\}.
\end{equation}
For a trajectory $\statetraj$, the outcome of a system execution in an environment $\env$, $\status$ returns $1$ if the trajectory satisfies the task specification, or a negative number if it does not; different numbers correspond to different types of failures (violations).

\subsection*{Illustration: an autonomous car}
\label{sec:running-example}

While the ideas presented here apply to any autonomous system, to ground the discussion, we consider an autonomous car on an obstructed track, as depicted in \cref{fig:car}, for illustration. In this environment, variables define the track shape (parameters) and the locations of obstacles (elements). 
Each obstacle is represented by a vector of two variables $[x,y]$, both specified relative to the track shape. The state $\state_t$ defines the car's position and heading,
 steering angle and speed. 
The system uses a neural controller trained to steer the car to the end of the track while avoiding collisions (the task), regardless of the track's shape and obstacle placement. In this use case, we distinguish failure types by the obstacle the car crashes into. Accordingly, the $\status$ predicate returns $1$ if the car's trajectory satisfies the task, and $-i$ if it crashes on or close to the $i$-th obstacle.


\noindent\textbf{Problem definition:} Given a failing execution (e.g., of an autonomous car), the analysis goal is to be able to \emph{reproduce} it, 
generating additional similar failing executions, and to \emph{avoid} it, introducing minimal changes to repair the failure.

\section{Actual Causality for Failure Explanation: Fundamentals}\label{sec:fundamentals}

In this section, we introduce the framework of actual
causality and our causal setup for failure explanation.

\subsection{Actual causality}
In this section, we provide a brief, simplified overview of actual causality. 
The reader is referred to~\cite{HP01b,Hal16} for the full exposition.

Actual causality~\cite{HP01b,Hal16} provides a formal framework for reasoning about the causes of observed outcomes. The framework describes the world in terms of variables and their values. A causal model $M$ consists of a set of variables $\V$ and a set of structural equations that capture dependencies between these variables. Some salient features of causal models can be represented by a \emph{causal network}: a directed acyclic graph whose nodes are the variables of $M$ and whose directed edges capture the existence and direction of causal dependence. 

A \emph{context} $\vec{u}$ is an assignment to the variables of $M$ that have no incoming edges in the causal network (that is, their values do not depend on the values of other variables). $\vec{u}$ determines, through the equations of $M$, the values of all variables in $\V$. 
Together, a causal model $M$ and a context $\vec{u}$, which specifies the particular situation being analyzed, form a causal setting $(M,\vec{u})$.
We write $(M,\vec{u}) \sat \varphi$ if a Boolean formula $\varphi$ holds in the causal setting $(M,\vec{u})$.

Causal analysis is performed through \emph{interventions}. An intervention $\vec{X}\leftarrow\vec{x}$ assigns values $\vec{x}$ to a subset $\vec{X}\subseteq\V$, replacing the corresponding equations with these assignments. We write $(M,\vec{u})\models[\vec{X}\leftarrow\vec{x}]\varphi$ if $\varphi$ holds after this intervention. Interventions therefore allow us to reason about the effect of changing selected variables on the outcome.

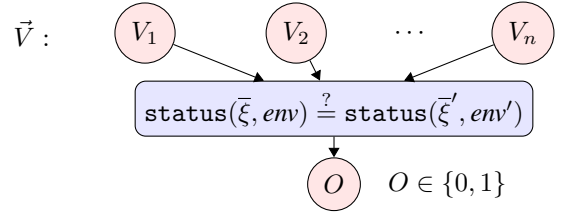
\begin{figure}[t]
\centering
    \begin{tikzpicture}[outer sep=auto]
    \node (V) at (-5, 0) {$\vec{V}:$};
    \node (v1) at (-3.5, 0) [draw, circle, rounded corners, minimum size=0.5cm,fill=red!10] {$V_1$};
    \node (v2) at (-1.5, 0) [draw, circle, rounded corners, minimum size=0.5cm,fill=red!10] {$V_2$};
    \node (dots) at (0, 0) {$\mathbf{\cdots}$};
    \node (vn) at (1.5, 0) [draw, circle, rounded corners, minimum size=0.6cm,fill=red!10] {$V_n$};

    \node (f) at (-1, -1.0) [draw, rectangle, rounded corners, fill=blue!10, minimum size=0.6cm]{$\status(\statetraj,\env) \? \status(\statetraj',\env')$};

    \draw [-Triangle] (v1) -- (f);
    \draw [-Triangle] (v2) -- (f);
    \draw [-Triangle] (vn) -- (f);
    
    \node (o) at (-1, -2.0) [draw, circle, minimum size=0.5cm, fill=red!10] {$O$};
    \node (oin) at (0.5, -2.0) {$O \in \{0,1\}$};

    \draw [-Triangle] (f) -- (o);
    \end{tikzpicture}
    \caption{A graphical representation of the causal model used to explain failures of our autonomous-car system. The red circles represent the variables in the causal model. The variable $O$ is determined from the simulation outcome, with the system treated as a black box.} 
    \label{fig:causal_model}
\end{figure}

\subsection{Causal framework for failure explanation}

We now instantiate this framework for explaining system failures, adapting some definitions from~\cite{icra} and restating them here. 
Given a crash scene $s=(\env,\state_0)$, we define a causal model $M_{s}$ as follows. The set $\V = \vec{V} \cup \{O\}$ of model variables consists of a set $\vec{V}$ corresponding to the obstacles in $\env$ and a single binary outcome variable $O$. Essentially, $\vec{V}$ is a Boolean \emph{mask} that indicates which obstacles are present and which are removed, allowing us to intervene on parts of $\env$. The outcome variable $O$ indicates whether the outcome of an execution under this masking matches that of the original scene. That is, 
$$O = [\status(\statetraj,\env)\?\status(\statetraj',\env')],$$ 
where $\env'$ is the environment induced by the (possibly intervened) causal setting, and $\statetraj'$ is its resulting trajectory. This results in a depth-$2$ causal model, graphically depicted in~\cref{fig:causal_model}, where each obstacle is (causally) independent of the others, meaning that each obstacle can be (re)moved without affecting the locations of other obstacles. Note that the causal model is specific for a particular scene $s$.
The controller itself is viewed as a black box, and we infer its behavior only from the simulation runs.


\setlength{\fboxsep}{1pt}  
\begin{figure*}[ht]
    \begin{subfigure}{0.325\linewidth}
        \centering
        \fbox{\includegraphics[scale=0.22,trim=23pt 30pt 23pt 21pt,clip]{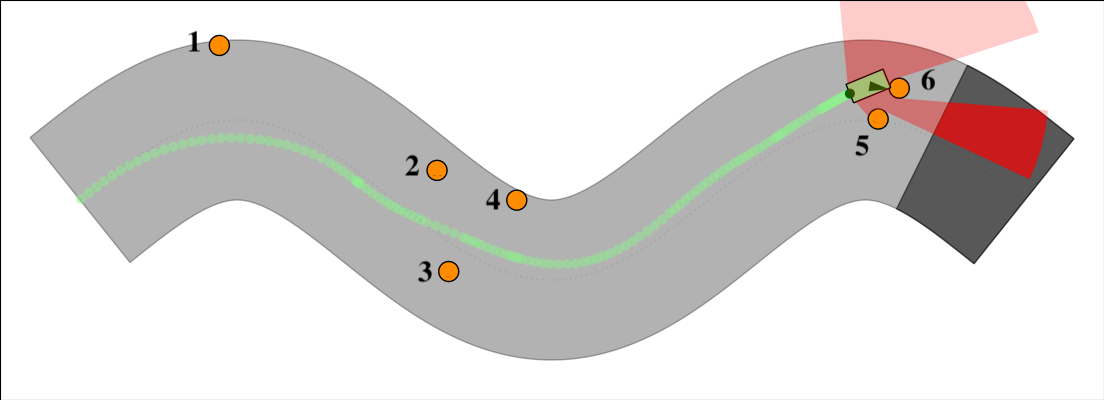}}
        \caption{Falsifying example}
        \label{fig:orig}
    \end{subfigure}
        \begin{subfigure}{0.325\linewidth}
        \centering
        \fbox{\includegraphics[scale=0.22,trim=23pt 30pt 23pt 21pt,clip]{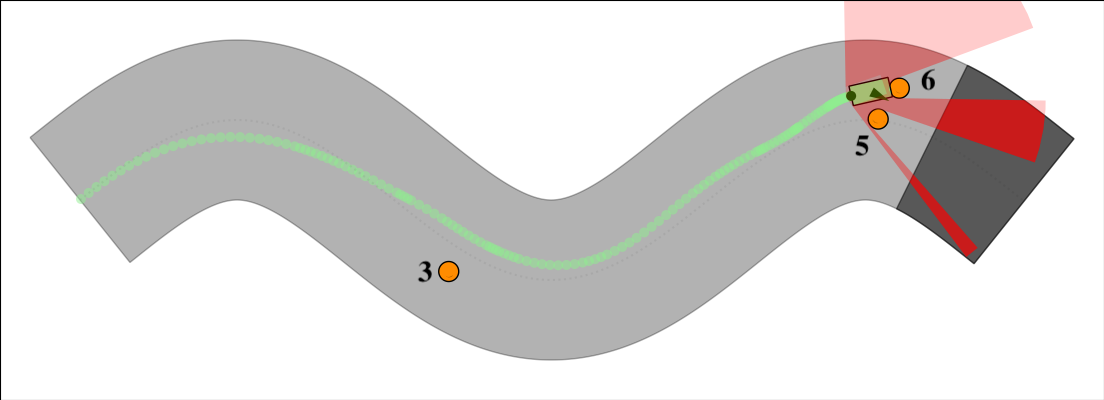}}
        \caption{Sufficient explanation}
        \label{fig:suf}
    \end{subfigure}
        \begin{subfigure}{0.325\linewidth}
        \centering
        \fbox{\includegraphics[scale=0.22,trim=23pt 30pt 23pt 21pt,clip]{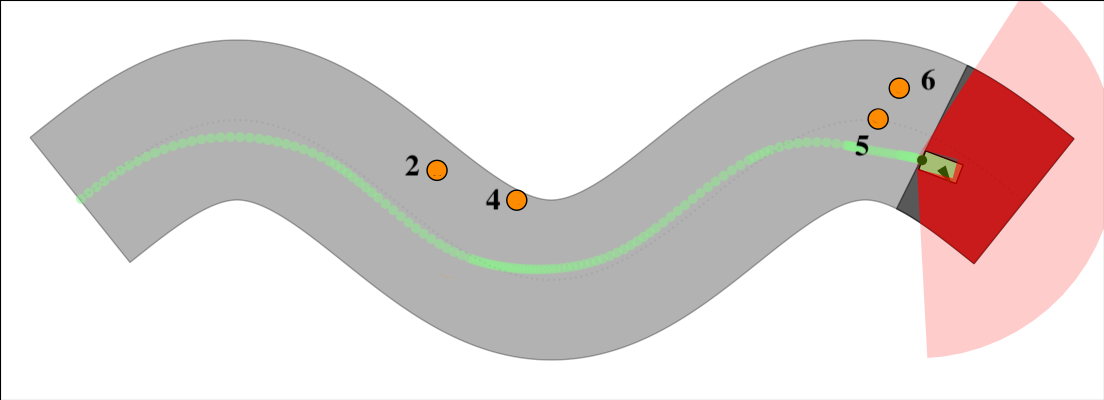}}
        \caption{Necessary explanation}
        \label{fig:nec}
    \end{subfigure}
    \caption{An illustration of the difference between sufficient and necessary explanations. 
    }
        \label{fig:nec_suf_example}
\end{figure*}

\subsection{Sufficient explanations for test outcomes}
Using the causal model $M_{s}$, we can formally characterize which scene elements caused the failure. \cite{icra} defines a sufficient explanation as follows, where $\vec{u}_1$ is the context that assigns $1$ to all variables in $\vec{V}$, and $\vec{u}_0$ is the context
that assigns $0$ to all these variables (in other words, $\vec{u}_1$ matches the original crash scene, and $\vec{u}_0$ matches the empty track).

\dfn[Sufficient Explanation~\cite{icra}]\label{defn:simple-exp}
A subset $\vec{V}_{\suf}$ of $\vec{V}$ is a \emph{sufficient explanation} of the outcome $O = 1$ 
if the following conditions hold:
\begin{description}
\item[{\rm EX1.}] $(M_{s},\vec{u}_0) \models [\vec{V}_{\suf} \leftarrow 1](O=1)$,
where $\vec{V}_{\suf} \leftarrow 1$ stands for assigning $1$ to all variables in 
$\vec{V}_{\suf}$.
\item[{\rm EX2.}] $\vec{V}_{\suf}$ is minimal; there is no strict subset $\vec{V}'_{\suf}$ of $\vec{V}_{\suf}$ that satisfies EX1, where $\vec{v}'$ is the restriction of $\vec{v}$ to the variables in $\vec{V}'_{\suf}$.
\end{description}
\edfn

Since there is a one-to-one correspondence between the variables in $\vec{V}$ and the obstacles in the scene, $\vec{V}_{\suf}$ is mapped one-to-one to a subset of obstacles that is sufficient for the observed crash. In other words, a sufficient explanation is a minimal subset of obstacles in a given scene $s$ that is \emph{sufficient} for the system to crash when others are removed. It is obtained by examining the result of setting $\vec{V}_{\suf}$ to $1$ in the context $\vec{u}_0$, that is, all obstacles except those in $\vec{V}_{\suf}$ are removed. Importantly, there can be more than one minimal sufficient explanation~\cite{CKK25}.

This definition of sufficient explanation addresses the task of failure understanding and reproduction. A sufficient explanation isolates a minimal sufficient set of conditions for the observed failure, thus helping us to understand why the failure occurred. 
Moreover, this minimal sufficient set can be used
to produce additional failure scenarios, supporting \emph{failure amplification}: by preserving the elements in the sufficient explanation while varying other elements of the environment, additional scenes exhibiting the same failure can be generated. This process is formal and structured, and it is particularly useful when failures are rare and obtaining additional failing executions is otherwise difficult.

\section{Identifying explanation objectives}
\label{sec:concept}



As a conceptual contribution, we identify two explanation objectives, distinguishing between \emph{outcome avoidance} and \emph{outcome reproduction}.
In the first case, the reason for explaining the test case is understanding what can/should be done in order to change the outcome. Considering failure outcomes, this is a repair-oriented objective. Such a repair can be initiated online, when the failure occurs (e.g., through human intervention), in order to allow the system to complete its task. Alternatively, it can be initiated after the fact to ensure the failure does not repeat.

In the latter case, the reason for explaining the test case is to understand how we can make this outcome happen again. This is especially important when a particular outcome is rare, and when one is encountered, we aim to maximize the lessons we learn from it, or repeat variations of it, to  ensure it is resolved. This can happen in later stages of system debugging or falsification, when looking for edge failure cases to robustify the system. We may also face it in the very early stages of development, when the positive outcome is rare, and we want to reinforce a desired behavior. 

Both types of explanations give us insight into the causes that led to the failure, but from complementary angles.
Notably, one is not automatically deducible from the other. Knowing how to reproduce a failure does not necessarily tell us how to fix it (and vice versa). The distinction is crucial. 

As we mentioned in the introduction, these two objectives coincide with the two main lines of work in the field: avoidance covers the line of work focusing on counterfactual generation, while reproduction covers sensitivity analysis.
As we show below, with the appropriate extension we contribute in this paper, the framework of actual causality allows us to create explanations that address both objectives.



\section{Actual Causality for Failure Explanation: Extension for Failure Avoidance}
\label{sec:ac-for-cps}

Aligned with the previous discussion, we now recognize that the concept of \emph{sufficient explanations}, as developed previously in \cite{icra}, naturally (only) addresses the objective of failure reproduction. 
This type of explanation, however, does not align with the objective of failure avoidance. 


Thus, we now extend the definitions presented in \cref{sec:fundamentals} and introduce the notion of necessary explanations.
This notion aligns naturally with the objective of failure avoidance.

\dfn[Necessary Explanation]\label{defn:necessary-exp}
A subset $\vec{V}_{\nec}$ of $\vec{V}$ is a \emph{necessary explanation} of the outcome, if the following conditions hold:
\begin{description}
\item[{\rm EX1.}] $(M_s,\vec{u}_1) \models [\vec{V}_{\nec} \leftarrow 0](O=0)$,
where $\vec{V}_{\nec} \leftarrow 0$ stands for assigning $0$ to all variables in 
$\vec{V}_{\nec}$.
\item[{\rm EX2.}] $\vec{V}_{\nec}$ is minimal; there is no strict subset $\vec{V}'_{\nec}$ of $\vec{V}_{\nec}$ that satisfies EX1, where $\vec{v}'$ is the restriction of $\vec{v}$ to the variables in $\vec{V}'_{\nec}$.
\end{description}
\edfn
In simple words, a necessary explanation is a minimal set of elements that, if removed, would cause the system to no longer fail in the same way.
Similarly to sufficient explanations, there can be multiple necessary explanations.

The key distinction between sufficient and necessary explanations lies in the direction of the intervention. A sufficient explanation starts from the empty scene, represented by $\vec{u}_0$, and identifies a minimal subset of elements whose presence is sufficient to reproduce the observed failure. In contrast, a necessary explanation starts from the original failing scene, represented by $\vec{u}_1$, and identifies a minimal subset of elements whose removal changes the outcome.

The distinction between necessity and sufficiency conditions is important in environments with intricate failure-inducing factors. For example, the removal of an obstacle subset may eliminate the failure, however, this subset alone may not be sufficient to cause the crash. Less intuitively, the removal of a subset sufficient to cause the failure may not eliminate the failure, as there can be multiple sufficiencies.

\Cref{fig:nec_suf_example} illustrates this distinction. In the running example, shown in \cref{fig:orig}, obstacles 3, 5, and 6 form a sufficient explanation because, by themselves, they are sufficient to reproduce the original crash (see \cref{fig:suf}), and no strict subset of them is. However, obstacle 3 is not necessary for the failure, since removing it alone from the original scene does not change the outcome (obstacles 1, 5, and 6 also form a sufficient explanation). In contrast, obstacles 1 and 3 form a necessary explanation (\cref{fig:nec}): removing them together changes the outcome, even though these obstacles alone are not sufficient to reproduce the original crash (note that removing obstacle 6, on which the vehicle crashes, clearly eliminates this particular crash, but this is not an interesting explanation).

In other words, there is a correspondence between the two types of explanations and two explanation objectives: sufficiency asks which elements are enough to reproduce the failure, whereas necessity asks which elements must be modified to avoid it.

A sufficient explanation can be used to produce additional examples of the same failure type (failed tests), by creating new failure-sufficient scenes---preserving the elements in the sufficient explanation while varying other elements---and re-running the system. 
A necessary explanation indicates what changes to the scene would change the (failed) outcome, while avoiding unnecessary changes. Thus, a necessary explanation can be used to generate, as a secondary artifact, high-quality \emph{counterfactuals}: execution examples of the system in a similar environment, in which the failure no longer occurs. Here, high-quality means that the difference between the counterfactual and the original scene is minimal, without unnecessary changes.

Overall, this extension allows us to cover both objectives of failure analysis within a unified causal formal framework, providing both a common language and comprehensive explanatory power.

\section{Generating necessary explanations} 
\label{sec:algorithm}


The precise computation of both types of explanations is intractable~\cite{CKKS26}. Consequently, existing causal explanation-generation approaches use heuristic search strategies~\cite{CKKS26}. These strategies are based on ranking the candidates according to their \emph{degree of responsibility}~\cite{CH04}, which 
\emph{quantifies} causal influence. In our setting, degree of responsibility is translated to the following. 
\dfn[Responsibility\cite{CH04,CKKS26}]\label{def:simple-resp}
The \emph{degree of responsibility} of $v_i$ for the outcome $O=o$ is defined as
$1/|\vec{V}_{\nec}|$, where $\vec{V}_{\nec} \subseteq \vec{V}$ is the \emph{smallest} necessary explanation for $O=o$ that 
contains $v_i$. 
If there is no such $\vec{V}_{\nec}$, then $v_i$'s degree of responsibility is defined as $0$.
\edfn
The precise computation of the degree of responsibility is, of course, also intractable~\cite{CKKS26}. However, \cite{CKKS26} introduces
an efficient algorithm for computing approximate degrees of responsibility and 
uses the values to guide the explanation construction.
This is also the strategy we follow here. 


We present an algorithm for deriving necessary explanations of system failures, in the context of our running example, based on \cref{defn:necessary-exp} and the causal model introduced in \Cref{sec:fundamentals}. The algorithm's structure is similar to the one introduced in~\cite{icra},
with both the construction and the search procedure modified to target necessary explanations. 
Essentially, we rank the obstacles according to their approximate degree of responsibility for the failure, and use this ranked list as a heuristic guidance in constructing the explanations.


\cref{alg:explanation} samples $m$ random interventions on repairable (movable) obstacles observed in the original trajectory $\statetraj$ (ln.~\ref{line:resp-start-1}--\ref{line:sample}) and evaluates their effect on the outcome. Among the outcome-changing interventions, only minimal subsets are retained, and their elements are assigned responsibility according to \cref{def:simple-resp} (ln.~\ref{line:resp-loop-start}--\ref{line:resp-end}).
The algorithm then ranks the obstacles by their estimated responsibility (ln.~\ref{line:ranking}) and greedily constructs explanation candidates using the ranking, removing obstacles from the original scene until it identifies a subset whose intervention changes the outcome (ln.~\ref{line:exp-outcome-changed}). This subset, by construction, is a necessary but not always minimal explanation. Hence, the algorithm minimizes the constructed explanation by removing redundant obstacles while ensuring that the remaining subset still changes the outcome, following \cref{defn:necessary-exp} (ln.~\ref{line:exp-loop-2-start}--\ref{line:exp-loop-2-end}).

\begin{algorithm}[t]
\footnotesize
\caption{Necessary explanation generation}
\label{alg:explanation}
\renewcommand{\algorithmicrequire}{\textbf{Input:}}
\begin{algorithmic}[1]
    \Require{Crash scene $s=(\env,\state_0)$, sampling budget $m$}
    \State{$elements \gets \text{element\_set\_in\_the\_environment($\env$)}$}
    \State{$\statetraj \gets \text{run\_simulation($s$)}$}
    \Statex{\Comment{Responsibility approximation of elements }}
    \State{$resp(e) \gets 0$ for every $e$ in $elements$}
    \State{\label{line:resp-start-1}$O \gets \text{observed\_repairable\_elements}(\statetraj, \env)$}
    \State{\label{line:sample}$\mathcal{I} \gets \text{sample\_m\_random\_interventions}(O, m)$}
    \State{\label{line:resp-start-2}$candidateSubsets \gets \emptyset$}
    \For{\label{line:resp-loop-start}$s'=(\env',\state_0)$ in $\mathcal{I}$}
        \State{$\statetraj' \gets \text{run\_simulation($s'$)}$}
        \If{$\status(\statetraj,\env) \ne \status(\statetraj',\env')$} \Comment{If outcome changed }
            \State{$S \gets \text{get\_intervened\_elements}(\env, \env')$}
            \State{\label{line:resp-loop-end}$candidateSubsets \gets candidateSubsets \cup S$}       
        \EndIf
    \EndFor
    \Statex{\Comment{Include only minimal subsets for responsibility computation}}
    \State{\label{line:resp-min}$minSubsets \gets \text{exclude\_nonminimal\_subsets($candidateSubsets$)}$}
    \For{\label{line:resp-loop-2-start}$S$ in $minSubsets$}
        \For{$e$ in $S$}
            \State{\label{line:resp-loop-2-end}\label{line:resp-end}$resp(e) \gets \text{max}(resp(e), 1/|S|)$ \Comment{Update responsibility (\cref{def:simple-resp})}}
        \EndFor
    \EndFor
    \Statex{\Comment{Responsibility-guided ranking of elements }}
    \State{\label{line:ranking}$el\_sorted \gets \text{sort\_by\_responsibility}(elements,resp)$}
    \For{\label{line:exp-loop}$i$ in $\{ 1, \dots , |el\_sorted| \}$}\Comment{Explanation candidates evaluation}
        \State{$S \gets el\_sorted[0:i] $}
        \State{$s' \gets \text{apply intervention to $s$ that removes all obstacles in $S$}$}
        \State{$\statetraj' \gets \text{run\_simulation($s'$)}$}
        \If{\label{line:exp-outcome-changed}$\status(\statetraj,\env) \ne \status(\statetraj',\env')$} \Comment{If outcome changed }
            \State{\label{line:exp-init}$V_{\nec} \gets \{ e \in elements \mid e \text{ in } el\_sorted[:i] \}$}
            \For{\label{line:exp-loop-2-start}$S \in 2^{V_{\nec}}$}\Comment{Finding a minimal explanation in $V_{\nec}$}
                \State{$s' \gets \text{apply intervention to $s$ that removes all obstacles in $S$}$}
                \State{$\statetraj' \gets \text{run\_simulation($s'$)}$}
                \If{$\status(\statetraj,\env) \ne \status(\statetraj',\env')$}   
                    \If{$|S| < |V_{\nec}|$}   
                        \State{\label{line:exp-loop-2-end}$V_{\nec} \gets S$}
                    \EndIf
                \EndIf
            \EndFor
            \State{\textbf{break}}
        \EndIf
    \EndFor
    \State{\Return{$V_{\nec} $}}

\end{algorithmic}
\end{algorithm}

\textbf{Guarantees:}
The algorithm operates under a reasonable assumption that a necessary explanation always exists. Indeed, removing all obstacles should eliminate
the crash (we ignore the degenerate cases where the track geometry itself causes the crash). 
Under this assumption the algorithm is guaranteed to return a necessary explanation, regardless of the accuracy of the responsibility approximation.
The responsibility only determines the ranking, thus in the worst case, all obstacles are considered.
In the following section, we empirically evaluate the effect of this responsibility-guided search on the computational effort required to find necessary explanations.

\section{Performance Evaluation}
\label{sec:performance-evaluation}

Since \cref{alg:explanation} is guaranteed to find a valid necessary explanation, we empirically validate its performance across different environments by measuring the computational effort required to obtain an explanation. The complete code for the evaluations and the implementation of \Cref{alg:explanation} are publicly available at~\cite{supplementarymaterial}. We ran the experiments on an Intel(R) Xeon(R) Platinum 8468 with 64GB RAM.

\begin{figure}[t]
\centering
\includegraphics[width=\columnwidth,height=3.6cm,trim=70pt 0pt 10pt 180pt,clip]{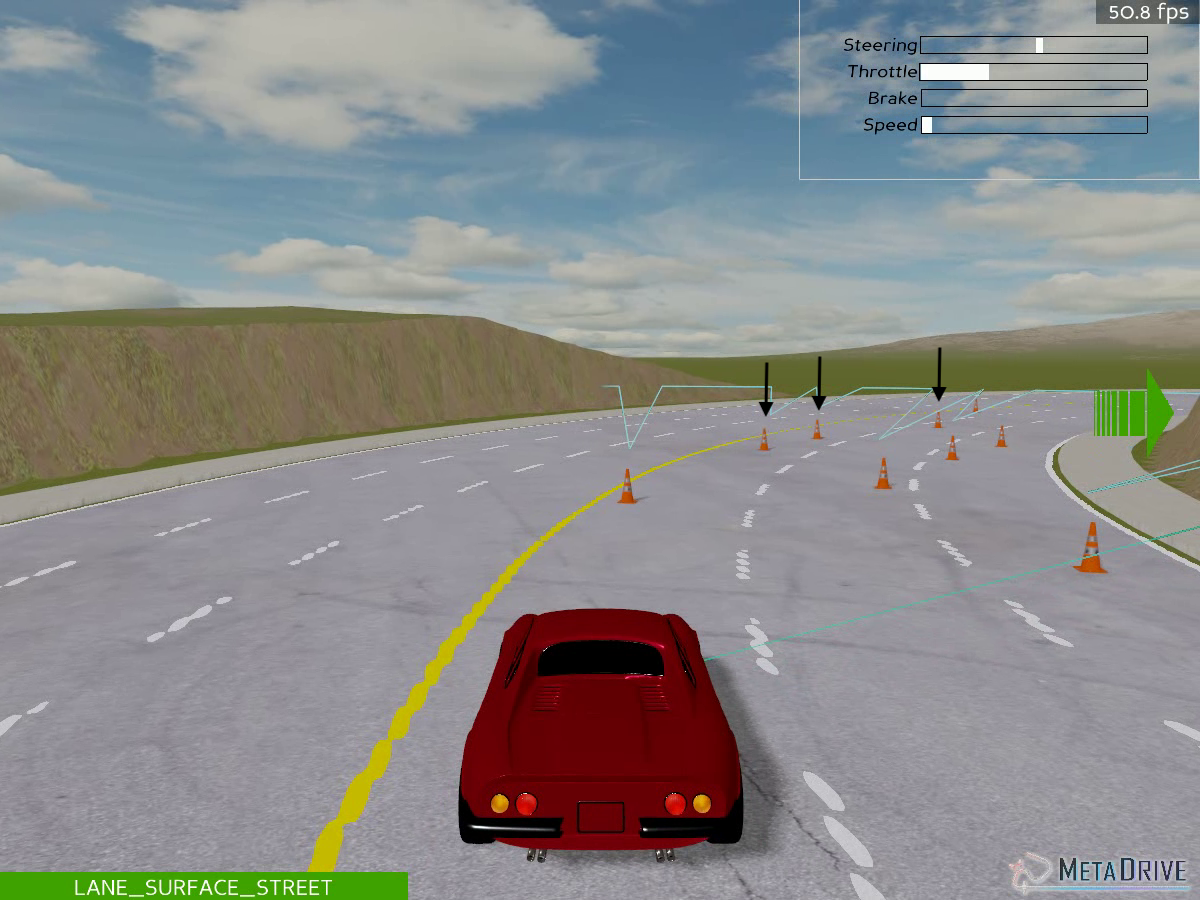}
\caption{The Metadrive simulation~\cite{li2022metadrive}. The arrows point to obstacles that form a necessary explanation.}
\label{fig:metadrive}
\end{figure}

\noindent\textbf{Experimental Setup.}
We evaluated our approach using two different autonomous-driving simulators: LiteRacer (LR)~\cite{Elimelech24LiteRacer} (see \cref{fig:car}) and MetaDrive~\cite{li2022metadrive} (see \cref{fig:metadrive}). In each simulator, the vehicle was controlled by its default neural network-based controller. To evaluate the generalizability of our approach across different environments, we considered several track geometries in each simulator. For LiteRacer, we constructed three track types: (1) \emph{Sinusoidal (Sin)}, following a sine-wave shape; (2) \emph{Segmented (Seg)}, consisting of both straight and curved segments; and (3) \emph{Lane Shift (LS)}, consisting primarily of straight segments connected by turns resembling lane shifts. Each track type was evaluated under four configurations combining short/long (S/L) tracks and narrow/wide (N/W) widths. For MetaDrive, we constructed three track types: (1) \emph{Curve}, (2) \emph{Straight}, and (3) \emph{Mixed}, which combines straight and curved segments. A visualization of the different tracks is available at~\cite{supplementarymaterial}.

For each track configuration, we generated $100$ failing examples with different obstacle placements,
with the number of obstacles from $15$ to $19$, with $20$ examples for each.
To ensure that additional obstacles indeed increase the complexity of the problem, we consider only examples in which all obstacles are observed by the vehicle during execution. Consequently, every obstacle could potentially influence the vehicle's behavior and is a candidate for inclusion in an explanation. The
obstacle on which the vehicle crashes is fixed, hence forcing the algorithm to look for other, more interesting, necessities.
This also models scenarios in which some environmental elements are unrepairable and therefore cannot be modified for failure avoidance.

\noindent\textbf{Comparison.}
We evaluated our method with a sample budget of $50$ to estimate causal responsibility.  
We also experimented with lower and higher budgets which yielded similar, though slightly worse, results.
To assess its efficiency, we compare our approach with the exhaustive search, which examines subsets of obstacles in increasing order of their size and terminates when it finds a subset that satisfies EX1 in \cref{defn:necessary-exp}. Since subsets are considered in increasing order of size, the resulting explanation also satisfies EX2. 

We measure computational effort by the total number of simulation runs performed by each method, as it dominates the computational cost in practice
(see the empirical analysis confirming this observation at~\cite{supplementarymaterial}). 

\begin{table}[t]
    \scriptsize
    \centering

    \caption{%
        Efficiency of explanation generation algorithms (number of simulation runs; lower is better)
    }
    \label{tab:runs-results}

    \scriptsize
    \setlength{\tabcolsep}{2.5pt}

    \begin{tabular}{c c || cc cc cc cc cc}

        \toprule
        \multicolumn{2}{c}{} &
        \multicolumn{2}{c}{\textbf{15}} &
        \multicolumn{2}{c}{\textbf{16}} &
        \multicolumn{2}{c}{\textbf{17}} &
        \multicolumn{2}{c}{\textbf{18}} &
        \multicolumn{2}{c}{\textbf{19}}
        \\
        \cmidrule(lr){3-4}\cmidrule(lr){5-6}\cmidrule(lr){7-8}\cmidrule(lr){9-10}\cmidrule(lr){11-12}
        & \textbf{Env.} &
        \textbf{Exh.} & \textbf{Resp.} &
        \textbf{Exh.} & \textbf{Resp.} &
        \textbf{Exh.} & \textbf{Resp.} &
        \textbf{Exh.} & \textbf{Resp.} &
        \textbf{Exh.} & \textbf{Resp.}
        \\
        \midrule


        \multirow{12}{*}{\rotatebox[origin=c]{90}{LiteRacer}}
        & Sin-SN
        & 442 & \textbf{161}
        & 877 & \textbf{559}
        & 1270 & \textbf{901}
        & 2584 & \textbf{1909}
        & 2816 & \textbf{2076}
        \\

        & Sin-SW
        & 224 & \textbf{93}
        & 240 & \textbf{114}
        & 298 & \textbf{140}
        & 394 & \textbf{142}
        & 386 & \textbf{171}
        \\

        & Sin-LN
        & 290 & \textbf{148}
        & 485 & \textbf{129}
        & 726 & \textbf{475}
        & 745 & \textbf{580}
        & 808 & \textbf{481}
        \\

        & Sin-LW
        & 208 & \textbf{94}
        & 232 & \textbf{56}
        & 339 & \textbf{132}
        & 328 & \textbf{158}
        & 383 & \textbf{176}
        \\
        \cmidrule(lr){2-12}

        & Seg-SN
        & 737 & \textbf{501}
        & 710 & \textbf{520}
        & 1050 & \textbf{690}
        & 1764 & \textbf{1126}
        & 3169 & \textbf{2305}
        \\

        & Seg-SW
        & 159 & \textbf{62}
        & 262 & \textbf{152}
        & 282 & \textbf{135}
        & 521 & \textbf{203}
        & 393 & \textbf{223}
        \\

        & Seg-LN
        & 340 & \textbf{185}
        & 324 & \textbf{157}
        & 678 & \textbf{301}
        & 447 & \textbf{97}
        & 1380 & \textbf{1041}
        \\

        & Seg-LW
        & 161 & \textbf{66}
        & 188 & \textbf{59}
        & 290 & \textbf{88}
        & 264 & \textbf{60}
        & 339 & \textbf{138}
        \\
        \cmidrule(lr){2-12}

        & LS-SN
        & 622 & \textbf{377}
        & 597 & \textbf{339}
        & 456 & \textbf{292}
        & 607 & \textbf{320}
        & 1631 & \textbf{1189}
        \\

        & LS-SW
        & 171 & \textbf{61}
        & 199 & \textbf{82}
        & 246 & \textbf{80}
        & 300 & \textbf{66}
        & 286 & \textbf{110}
        \\

        & LS-LN
        & 361 & \textbf{226}
        & 314 & \textbf{119}
        & 430 & \textbf{269}
        & 609 & \textbf{360}
        & 1437 & \textbf{875}
        \\

        & LS-LW
        & 161 & \textbf{52}
        & 217 & \textbf{66}
        & 243 & \textbf{111}
        & 301 & \textbf{76}
        & 402 & \textbf{161}
        \\
        \cmidrule(lr){2-12}


        \multirow{3}{*}{\rotatebox[origin=c]{90}{MetaDr.}}
        & Curve
        & 437 & \textbf{343}
        & 448 & \textbf{377}
        & 730 & \textbf{579}
        & 652 & \textbf{568}
        & 340 & \textbf{324}
        \\

        & Straight
        & 761 & \textbf{573}
        & 475 & \textbf{413}
        & 1601 & \textbf{1213}
        & 822 & \textbf{685}
        & 972 & \textbf{836}
        \\

        & Mixed
        & 284 & \textbf{255}
        & 1050 & \textbf{759}
        & 508 & \textbf{454}
        & 2047 & \textbf{1607}
        & 799 & \textbf{713}
        \\
        \bottomrule
    \end{tabular}

\end{table}

\noindent\textbf{Results.}
\Cref{tab:runs-results} reports the number of simulation runs of the exhaustive and responsibility-guided approaches. The responsibility-guided approach requires fewer simulation runs than exhaustive search across all evaluated environments and obstacle counts. The magnitude of the improvement 
is more pronounced in LiteRacer.
In some MetaDrive configurations, the guidance provided by the responsibility ranking appears insufficient to fully offset the additional simulation budget required to estimate it. Overall, these results show that approximate causal responsibility reduces the search effort required to obtain a necessary explanation, although the extent of this reduction depends on the analyzed scenario.



\section{Counterfactual Generation Evaluation}
\label{sec:counterfactuals-evaluation}

We now evaluate the practical utility of our approach for failure avoidance. The task in this section is, given a failing example, to generate a single counterfactual: a similar example in which the system passes. A necessary explanation identifies a minimal subset of elements whose modification changes the outcome of a failing execution. It can therefore be viewed as a targeted counterfactual search: rather than searching directly for a passing scenario, we identify the elements that need to be modified and restrict the counterfactual search to those elements. The experimental setup is similar to that described in \cref{sec:performance-evaluation}.

\noindent\textbf{Comparison.}
Following previous work on counterfactual generation~\cite{guidotti2024counterfactual}, we evaluate the generated counterfactuals along three dimensions: 1) \emph{validity (correctness)}, measuring whether the counterfactual avoids the failure observed in the original example, 2) \emph{similarity}, measuring how close the counterfactual is to the original failing example, and 3) \emph{efficiency}, measuring the time required to generate the counterfactual. We quantify similarity using the Euclidean distance between the original and counterfactual examples.

For counterfactual generation, we implemented a process that uses a third of the simulation runs budget to identify necessary explanations using the algorithm described in~\cref{sec:algorithm}. It uses the remaining budget to efficiently perturb the failing example. Specifically, for each sample, the algorithm first randomly selects an explanation from the set of necessary explanations found in the first step. It then randomly perturbs elements from this set, leaving the rest unchanged.

For comparison, we evaluated our approach against two methods. The first is DeCaF, the counterfactual generation approach proposed in~\cite{ghazal2026towards}. DeCaF learns a model of system behavior from previously simulated passing and failing examples and uses it to guide the search for a passing example close to the original failure. Our analysis used the variation that achieved the best success rate according to~\cite{ghazal2026towards}, which uses an M5 model tree for model learning and a KD-Tree Nearest Neighbors search. The second method is a random-search baseline that samples perturbations of the original failing example. The returned example is the closest one that avoids the observed failure. We evaluated all algorithms under simulation run budgets of $100$, $200$, $300$, and $400$.

\noindent\textbf{Results.}
\cref{tab:average-accuracy} reports the counterfactual validity rate under a budget of $400$ simulation runs, measured as the percentage of cases in which each method successfully generated a failure-avoiding example. In LiteRacer environments, both our necessary explanation-based approach and the random baseline achieve a $100\%$ validity rate across all settings, whereas DeCaF fails to generate valid counterfactuals in several cases. The more challenging MetaDrive environments reveal clearer differences between the methods: although none achieves perfect success across all settings, our approach consistently achieves the highest validity rate, generating valid counterfactuals in the majority of cases and outperforming both methods across all settings. See~\cite{supplementarymaterial} for the results for the budgets of $100$, $200$, and $300$.

\begin{table}[t]
    \scriptsize
    \centering

    \caption{%
        Counterfactual validity (correctness) rate for the necessary explanations (our approach),
        DeCaF, and random methods with a sample budget of $400$. 
    }
    \label{tab:average-accuracy}

    \scriptsize
    \setlength{\tabcolsep}{1.6pt}

    \begin{tabular}{c c || ccc ccc ccc ccc ccc}

        \toprule
        \multicolumn{2}{c}{} &
        \multicolumn{3}{c}{\textbf{15}} &
        \multicolumn{3}{c}{\textbf{16}} &
        \multicolumn{3}{c}{\textbf{17}} &
        \multicolumn{3}{c}{\textbf{18}} &
        \multicolumn{3}{c}{\textbf{19}}
        \\
        \cmidrule(lr){3-5}
        \cmidrule(lr){6-8}
        \cmidrule(lr){9-11}
        \cmidrule(lr){12-14}
        \cmidrule(lr){15-17}

        & \textbf{Env.} &
        \textbf{NE} & \textbf{De} & \textbf{Ra} &
        \textbf{NE} & \textbf{De} & \textbf{Ra} &
        \textbf{NE} & \textbf{De} & \textbf{Ra} &
        \textbf{NE} & \textbf{De} & \textbf{Ra} &
        \textbf{NE} & \textbf{De} & \textbf{Ra}
        \\
        \midrule


        \multirow{12}{*}{\rotatebox[origin=c]{90}{LiteRacer}}
        & Sin-SN
        & \textbf{1.0} & .85 & \textbf{1.0}
        & \textbf{1.0} & \textbf{1.0} & \textbf{1.0}
        & \textbf{1.0} & \textbf{1.0} & \textbf{1.0}
        & \textbf{1.0} & .95 & \textbf{1.0}
        & \textbf{1.0} & \textbf{1.0} & \textbf{1.0}
        \\

        & Sin-SW
        & \textbf{1.0} & \textbf{1.0} & \textbf{1.0}
        & \textbf{1.0} & \textbf{1.0} & \textbf{1.0}
        & \textbf{1.0} & .95 & \textbf{1.0}
        & \textbf{1.0} & .90 & \textbf{1.0}
        & \textbf{1.0} & \textbf{1.0} & \textbf{1.0}
        \\

        & Sin-LN
        & \textbf{1.0} & \textbf{1.0} & \textbf{1.0}
        & \textbf{1.0} & .95 & \textbf{1.0}
        & \textbf{1.0} & .85 & \textbf{1.0}
        & \textbf{1.0} & \textbf{1.0} & \textbf{1.0}
        & \textbf{1.0} & \textbf{1.0} & \textbf{1.0}
        \\

        & Sin-LW
        & \textbf{1.0} & \textbf{1.0} & \textbf{1.0}
        & \textbf{1.0} & .90 & \textbf{1.0}
        & \textbf{1.0} & .90 & \textbf{1.0}
        & \textbf{1.0} & .90 & \textbf{1.0}
        & \textbf{1.0} & \textbf{1.0} & \textbf{1.0}
        \\
        \cmidrule(lr){2-17}

        & Seg-SN
        & \textbf{1.0} & .95 & \textbf{1.0}
        & \textbf{1.0} & .95 & \textbf{1.0}
        & \textbf{1.0} & .95 & \textbf{1.0}
        & \textbf{1.0} & .90 & \textbf{1.0}
        & \textbf{1.0} & .95 & \textbf{1.0}
        \\

        & Seg-SW
        & \textbf{1.0} & .95 & \textbf{1.0}
        & \textbf{1.0} & \textbf{1.0} & \textbf{1.0}
        & \textbf{1.0} & \textbf{1.0} & \textbf{1.0}
        & \textbf{1.0} & .95 & \textbf{1.0}
        & \textbf{1.0} & .95 & \textbf{1.0}
        \\

        & Seg-LN
        & \textbf{1.0} & .80 & \textbf{1.0}
        & \textbf{1.0} & .95 & \textbf{1.0}
        & \textbf{1.0} & .95 & \textbf{1.0}
        & \textbf{1.0} & .95 & \textbf{1.0}
        & \textbf{1.0} & .90 & \textbf{1.0}
        \\

        & Seg-LW
        & \textbf{1.0} & .95 & \textbf{1.0}
        & \textbf{1.0} & .80 & \textbf{1.0}
        & \textbf{1.0} & .90 & \textbf{1.0}
        & \textbf{1.0} & .90 & \textbf{1.0}
        & \textbf{1.0} & .95 & \textbf{1.0}
        \\
        \cmidrule(lr){2-17}

        & LS-SN
        & \textbf{1.0} & .90 & \textbf{1.0}
        & \textbf{1.0} & .95 & \textbf{1.0}
        & \textbf{1.0} & \textbf{1.0} & \textbf{1.0}
        & \textbf{1.0} & .95 & \textbf{1.0}
        & \textbf{1.0} & .90 & \textbf{1.0}
        \\

        & LS-SW
        & \textbf{1.0} & .95 & \textbf{1.0}
        & \textbf{1.0} & .95 & \textbf{1.0}
        & \textbf{1.0} & .95 & \textbf{1.0}
        & \textbf{1.0} & .90 & \textbf{1.0}
        & \textbf{1.0} & .95 & \textbf{1.0}
        \\

        & LS-LN
        & \textbf{1.0} & .85 & \textbf{1.0}
        & \textbf{1.0} & \textbf{1.0} & \textbf{1.0}
        & \textbf{1.0} & \textbf{1.0} & \textbf{1.0}
        & \textbf{1.0} & \textbf{1.0} & \textbf{1.0}
        & \textbf{1.0} & .85 & \textbf{1.0}
        \\

        & LS-LW
        & \textbf{1.0} & \textbf{1.0} & \textbf{1.0}
        & \textbf{1.0} & .95 & \textbf{1.0}
        & \textbf{1.0} & .70 & \textbf{1.0}
        & \textbf{1.0} & .95 & \textbf{1.0}
        & \textbf{1.0} & .90 & \textbf{1.0}
        \\
        \cmidrule(lr){2-17}


        \multirow{3}{*}{\rotatebox[origin=c]{90}{MetaDr.}}
        & Curve
        & \textbf{.85} & .60 & .70
        & \textbf{.95} & .80 & .85
        & \textbf{1.0} & .85 & .95
        & \textbf{.90} & .70 & \textbf{.90}
        & \textbf{1.0} & .75 & .90
        \\

        & Straight
        & \textbf{.75} & .40 & .45
        & \textbf{.90} & .45 & .55
        & \textbf{.80} & .65 & .55
        & \textbf{.75} & .40 & .60
        & \textbf{.85} & .35 & .40
        \\

        & Mixed
        & \textbf{.75} & .65 & .70
        & \textbf{.90} & .75 & .75
        & \textbf{1.0} & .75 & .95
        & \textbf{.95} & .75 & .85
        & \textbf{.95} & .45 & .80
        \\
        \bottomrule

    \end{tabular}

\end{table}

\cref{tab:average-distance} presents the average distance of the generated counterfactuals under a budget of $400$ simulation runs. To ensure a fair comparison, we restrict the analysis to examples where all three methods successfully generate a valid counterfactual, ensuring that the reported distances are computed over a similar, valid evaluation set. The results reveal a clear and consistent ordering among the methods. The necessary explanation-based approach generates the closest counterfactuals in almost every setting, outperforming both methods, and DeCaF generally ranks second. These results indicate that necessary explanations guide the search toward more targeted modifications. \cref{fig:counterfactual-distance} shows the average counterfactual distance as a function of the simulation-run budget for environments with 19 obstacles. While the necessary explanation-based approach and DeCaF achieve comparable performance at lower budgets, our approach increasingly outperforms DeCaF as the sample budget grows. Moreover, our approach exhibits more stable improvement, with counterfactual distance decreasing monotonically with the number of samples in almost all settings. In contrast, DeCaF shows several cases where increasing the sample budget leads to worse counterfactual quality. The results for settings with $15$--$18$ obstacles are provided at~\cite{supplementarymaterial}.

\begin{table}[t]
    \scriptsize
    \centering

    \caption{%
    Quality of counterfactuals generated using necessary explanations,
        DeCaF, and random methods (average counterfactual distance; lower is better).
    }
    \label{tab:average-distance}

    \scriptsize
    \setlength{\tabcolsep}{1.6pt}

    \begin{tabular}{c c || ccc ccc ccc ccc ccc}

        \toprule
        \multicolumn{2}{c}{} &
        \multicolumn{3}{c}{\textbf{15}} &
        \multicolumn{3}{c}{\textbf{16}} &
        \multicolumn{3}{c}{\textbf{17}} &
        \multicolumn{3}{c}{\textbf{18}} &
        \multicolumn{3}{c}{\textbf{19}}
        \\
        \cmidrule(lr){3-5}
        \cmidrule(lr){6-8}
        \cmidrule(lr){9-11}
        \cmidrule(lr){12-14}
        \cmidrule(lr){15-17}

        & \textbf{Env.} &
        \textbf{NE} & \textbf{De} & \textbf{Ra} &
        \textbf{NE} & \textbf{De} & \textbf{Ra} &
        \textbf{NE} & \textbf{De} & \textbf{Ra} &
        \textbf{NE} & \textbf{De} & \textbf{Ra} &
        \textbf{NE} & \textbf{De} & \textbf{Ra}
        \\
        \midrule


        \multirow{12}{*}{\rotatebox[origin=c]{90}{LiteRacer}}
        & Sin-SN
        & \textbf{.25} & .38 & .58
        & \textbf{.14} & .27 & .49
        & \textbf{.29} & .40 & .88
        & \textbf{.16} & .51 & .61
        & \textbf{.42} & .67 & 1.03
        \\

        & Sin-SW
        & \textbf{.15} & .18 & .31
        & \textbf{.15} & .24 & .43
        & \textbf{.16} & .25 & .39
        & \textbf{.22} & .40 & .56
        & \textbf{.25} & .33 & .49
        \\

        & Sin-LN
        & \textbf{.18} & .38 & .62
        & \textbf{.15} & .42 & .73
        & \textbf{.30} & .53 & .79
        & \textbf{.28} & .52 & .74
        & \textbf{.34} & .62 & 1.43
        \\

        & Sin-LW
        & \textbf{.13} & .28 & .40
        & \textbf{.18} & .31 & .57
        & \textbf{.26} & .32 & .62
        & \textbf{.13} & .22 & .65
        & \textbf{.23} & .38 & .80
        \\
        \cmidrule(lr){2-17}

        & Seg-SN
        & \textbf{.28} & .54 & .76
        & \textbf{.25} & .35 & .72
        & \textbf{.25} & .44 & .82
        & \textbf{.35} & .59 & .94
        & \textbf{.48} & .74 & 1.07
        \\

        & Seg-SW
        & \textbf{.10} & .19 & .36
        & \textbf{.14} & .23 & .36
        & \textbf{.19} & .29 & .61
        & \textbf{.15} & .24 & .31
        & \textbf{.19} & .31 & .49
        \\

        & Seg-LN
        & \textbf{.25} & .29 & .76
        & \textbf{.21} & .53 & .70
        & \textbf{.37} & .70 & .88
        & \textbf{.28} & .45 & .88
        & \textbf{.32} & .45 & 1.40
        \\

        & Seg-LW
        & \textbf{.17} & .26 & .45
        & \textbf{.20} & .25 & .53
        & \textbf{.17} & .31 & .72
        & \textbf{.22} & .33 & .65
        & \textbf{.21} & .45 & .92
        \\
        \cmidrule(lr){2-17}

        & LS-SN
        & \textbf{.11} & .23 & .42
        & \textbf{.12} & .27 & .46
        & \textbf{.27} & .50 & .67
        & \textbf{.21} & .30 & .82
        & \textbf{.47} & .63 & .89
        \\

        & LS-SW
        & \textbf{.17} & .21 & .37
        & \textbf{.13} & .26 & .24
        & \textbf{.15} & .22 & .36
        & \textbf{.20} & .34 & .57
        & \textbf{.16} & .28 & .46
        \\

        & LS-LN
        & .49 & \textbf{.39} & .70
        & \textbf{.18} & .51 & .73
        & \textbf{.20} & .54 & .73
        & \textbf{.31} & .40 & .97
        & \textbf{.34} & .50 & 1.14
        \\

        & LS-LW
        & \textbf{.12} & .25 & .39
        & \textbf{.19} & .24 & .43
        & \textbf{.21} & .27 & .55
        & \textbf{.13} & .26 & .51
        & \textbf{.30} & .38 & .71
        \\
        \cmidrule(lr){2-17}


        \multirow{3}{*}{\rotatebox[origin=c]{90}{MetaDr.}}
        & Curve
        & \textbf{.18} & .32 & .35
        & \textbf{.13} & .17 & .22
        & \textbf{.14} & .23 & .24
        & \textbf{.23} & .37 & .40
        & \textbf{.19} & .21 & .36
        \\

        & Straight
        & \textbf{.12} & .15 & .19
        & \textbf{.26} & .33 & .50
        & \textbf{.14} & .24 & .41
        & \textbf{.14} & .29 & .43
        & \textbf{.21} & .39 & .42
        \\

        & Mixed
        & \textbf{.14} & .18 & .23
        & \textbf{.15} & .22 & .28
        & \textbf{.17} & .24 & .39
        & \textbf{.13} & .22 & .32
        & \textbf{.13} & .22 & .29
        \\
        \bottomrule

    \end{tabular}

\end{table}

\pgfplotsset{
    necessaryDistance/.style={
        color=blue!80,
        very thick,
        mark=*,
        mark size=1.3pt
    },
    decafDistance/.style={
        color=red!80,
        thick,
        mark=square*,
        mark size=1.3pt
    },
    randomDistance/.style={
        color=black!70,
        thick,
        mark=triangle*,
        mark size=1.5pt
    }
}

\newcommand{\plotdistance}[1]{%
    \addplot[necessaryDistance]
        table[
            x=samples,
            y=necessary_explanations,
            col sep=comma
        ] {#1};

    \addplot[decafDistance]
        table[
            x=samples,
            y=decaf,
            col sep=comma
        ] {#1};

    \addplot[randomDistance]
        table[
            x=samples,
            y=random,
            col sep=comma
        ] {#1};
}

\newcommand{\plotdistancewithlegend}[1]{%
    \addplot[necessaryDistance]
        table[
            x=samples,
            y=necessary_explanations,
            col sep=comma
        ] {#1};
    \addlegendentry{Necessary Explanations}

    \addplot[decafDistance]
        table[
            x=samples,
            y=decaf,
            col sep=comma
        ] {#1};
    \addlegendentry{DeCaF}

    \addplot[randomDistance]
        table[
            x=samples,
            y=random,
            col sep=comma
        ] {#1};
    \addlegendentry{Random}
}

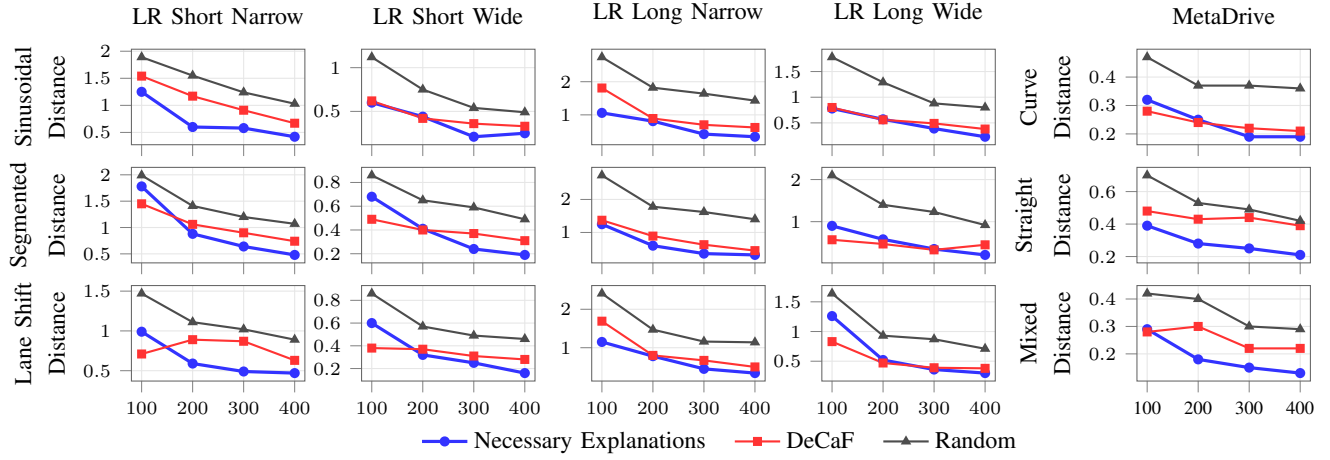
\begin{figure*}[t]
    \centering

    \begin{tikzpicture}

        \begin{groupplot}[
            group style={
                group name=plots,
                group size=6 by 3,
                horizontal sep=7.5mm,
                vertical sep=3mm,
            },
            width=0.218\textwidth,
            height=0.16\textwidth,
            xmin=80,
            xmax=420,
            xtick={100,200,300,400},
            scaled y ticks=false,
            tick align=outside,
            tick pos=left,
            grid=major,
            grid style={black!10},
            axis line style={black!60},
            tick label style={font=\scriptsize},
            label style={font=\small},
            title style={font=\small},
        ]


        \nextgroupplot[
            title={LR Short Narrow},
            ylabel={Distance},
            xticklabels=\empty,
            legend columns=5,
            legend to name=commonlegend,
            legend style={
                font=\small,
                draw=none,
                /tikz/every even column/.append style={
                    column sep=3mm
                }
            }
        ]
        \plotdistancewithlegend{results/sinusoidal_short_narrow_average_distance_obs_19.csv}

        \nextgroupplot[
            title={LR Short Wide},
            xticklabels=\empty,
        ]
        \plotdistance{results/sinusoidal_short_wide_average_distance_obs_19.csv}

        \nextgroupplot[
            title={LR Long Narrow},
            xticklabels=\empty,
        ]
        \plotdistance{results/sinusoidal_long_narrow_average_distance_obs_19.csv}

        \nextgroupplot[
            title={LR Long Wide},
            xticklabels=\empty,
        ]
        \plotdistance{results/sinusoidal_long_wide_average_distance_obs_19.csv}

        \nextgroupplot[
            hide axis,
            width=0.11\textwidth
        ]

        \nextgroupplot[
            title={MetaDrive},
            ylabel={Distance},
            xticklabels=\empty
        ]
        \plotdistance{results/metadrive_curve_average_distance_obs_19.csv}


        \nextgroupplot[
            ylabel={Distance},
            xticklabels=\empty
        ]
        \plotdistance{results/segmented_short_narrow_average_distance_obs_19.csv}

        \nextgroupplot[
            xticklabels=\empty,
        ]
        \plotdistance{results/segmented_short_wide_average_distance_obs_19.csv}

        \nextgroupplot[
            xticklabels=\empty,
        ]
        \plotdistance{results/segmented_long_narrow_average_distance_obs_19.csv}

        \nextgroupplot[
            xticklabels=\empty,
        ]
        \plotdistance{results/segmented_long_wide_average_distance_obs_19.csv}

        \nextgroupplot[
            hide axis,
            width=0.11\textwidth
        ]

        \nextgroupplot[
            ylabel={Distance},
            xticklabels=\empty
        ]
        \plotdistance{results/metadrive_straight_average_distance_obs_19.csv}


        \nextgroupplot[
            ylabel={Distance}
        ]
        \plotdistance{results/lane_shift_short_narrow_average_distance_obs_19.csv}

        \nextgroupplot[
        ]
        \plotdistance{results/lane_shift_short_wide_average_distance_obs_19.csv}

        \nextgroupplot[
        ]
        \plotdistance{results/lane_shift_long_narrow_average_distance_obs_19.csv}

        \nextgroupplot[
        ]
        \plotdistance{results/lane_shift_long_wide_average_distance_obs_19.csv}

        \nextgroupplot[
            hide axis,
            width=0.11\textwidth
        ]

        \nextgroupplot[
            ylabel={Distance},
            xtick={100,200,300,400},
            xticklabels={100,200,300,400}
        ]
        \plotdistance{results/metadrive_mixed_average_distance_obs_19.csv}

        \end{groupplot}





        \node[
            rotate=90,
            anchor=south,
            font=\small
        ] at ([xshift=-12mm]$(plots c1r1.west)$)
        {Sinusoidal};

        \node[
            rotate=90,
            anchor=south,
            font=\small
        ] at ([xshift=-12mm]$(plots c1r2.west)$)
        {Segmented};

        \node[
            rotate=90,
            anchor=south,
            font=\small
        ] at ([xshift=-12mm]$(plots c1r3.west)$)
        {Lane Shift};


        \node[
            rotate=90,
            anchor=south,
            font=\small
        ] at ([xshift=-12mm]$(plots c6r1.west)$)
        {Curve};

        \node[
            rotate=90,
            anchor=south,
            font=\small
        ] at ([xshift=-12mm]$(plots c6r2.west)$)
        {Straight};

        \node[
            rotate=90,
            anchor=south,
            font=\small
        ] at ([xshift=-12mm]$(plots c6r3.west)$)
        {Mixed};


        \node[
            anchor=north
        ] at ([yshift=-4mm]
             $(plots c1r3.south)!0.5!(plots c6r3.south)$)
        {\pgfplotslegendfromname{commonlegend}};

    \end{tikzpicture}

    \caption{%
        Quality of counterfactuals as a function of the number of simulation runs ($100$--$400$) for necessary explanations (our approach), DeCaF, and random methods, measured by the average distance from the original example (lower is better). 
    }
    \label{fig:counterfactual-distance}

\end{figure*}

\cref{tab:elapsed-time} 
shows the average elapsed time for each method with a budget of $400$ simulation runs. The necessary explanation-based approach exhibits runtimes comparable to the random baseline across both simulators, while DeCaF is consistently slower across all environments and obstacle counts. Importantly, this shows that the improved validity rate and counterfactual quality achieved by our approach do not come at the cost of increased execution time. 

\begin{table}[t]
    \scriptsize
    \centering

    \caption{%
        Elapsed time (minutes) for the necessary explanations,
        DeCaF, and random methods. 
    }
    \label{tab:elapsed-time}

    \scriptsize
    \setlength{\tabcolsep}{1.6pt}

    \begin{tabular}{c c || ccc ccc ccc ccc ccc}

        \toprule
        \multicolumn{2}{c}{} &
        \multicolumn{3}{c}{\textbf{15}} &
        \multicolumn{3}{c}{\textbf{16}} &
        \multicolumn{3}{c}{\textbf{17}} &
        \multicolumn{3}{c}{\textbf{18}} &
        \multicolumn{3}{c}{\textbf{19}}
        \\
        \cmidrule(lr){3-5}
        \cmidrule(lr){6-8}
        \cmidrule(lr){9-11}
        \cmidrule(lr){12-14}
        \cmidrule(lr){15-17}

        & \textbf{Env.} &
        \textbf{NE} & \textbf{De} & \textbf{Ra} &
        \textbf{NE} & \textbf{De} & \textbf{Ra} &
        \textbf{NE} & \textbf{De} & \textbf{Ra} &
        \textbf{NE} & \textbf{De} & \textbf{Ra} &
        \textbf{NE} & \textbf{De} & \textbf{Ra}
        \\
        \midrule


        \multirow{12}{*}{\rotatebox[origin=c]{90}{LiteRacer}}
        & Sin-SN
        & \textbf{11} & 18 & \textbf{11}
        & 10 & 16 & \textbf{9}
        & \textbf{9} & 17 & \textbf{9}
        & 9 & 15 & \textbf{8}
        & \textbf{8} & 15 & \textbf{8}
        \\

        & Sin-SW
        & \textbf{11} & 20 & \textbf{11}
        & \textbf{10} & 17 & \textbf{10}
        & 10 & 16 & \textbf{9}
        & \textbf{8} & 14 & \textbf{8}
        & \textbf{9} & 15 & \textbf{9}
        \\

        & Sin-LN
        & 19 & 27 & \textbf{18}
        & 16 & 23 & \textbf{15}
        & \textbf{14} & 20 & \textbf{14}
        & 15 & 23 & \textbf{14}
        & 15 & 22 & \textbf{14}
        \\

        & Sin-LW
        & \textbf{19} & 27 & \textbf{19}
        & \textbf{16} & 22 & \textbf{16}
        & \textbf{13} & 19 & 14
        & \textbf{15} & 21 & \textbf{15}
        & \textbf{15} & 23 & \textbf{15}
        \\
        \cmidrule(lr){2-17}

        & Seg-SN
        & 16 & 25 & \textbf{15}
        & 12 & 20 & \textbf{11}
        & 10 & 15 & \textbf{9}
        & \textbf{9} & 16 & \textbf{9}
        & 11 & 19 & \textbf{10}
        \\

        & Seg-SW
        & \textbf{16} & 24 & \textbf{16}
        & \textbf{13} & 20 & \textbf{13}
        & \textbf{10} & 16 & \textbf{10}
        & \textbf{10} & 15 & \textbf{10}
        & \textbf{12} & 18 & \textbf{12}
        \\

        & Seg-LN
        & 27 & 34 & \textbf{26}
        & \textbf{18} & 24 & \textbf{18}
        & 17 & 21 & \textbf{15}
        & 20 & 26 & \textbf{19}
        & \textbf{19} & 25 & \textbf{19}
        \\

        & Seg-LW
        & \textbf{27} & 34 & \textbf{27}
        & \textbf{18} & 25 & \textbf{18}
        & \textbf{17} & 23 & \textbf{17}
        & \textbf{20} & 28 & \textbf{20}
        & \textbf{19} & 25 & \textbf{19}
        \\
        \cmidrule(lr){2-17}

        & LS-SN
        & \textbf{10} & 16 & \textbf{10}
        & 10 & 17 & \textbf{9}
        & 10 & 17 & \textbf{9}
        & 9 & 16 & \textbf{8}
        & 10 & 16 & \textbf{9}
        \\

        & LS-SW
        & \textbf{10} & 16 & \textbf{10}
        & \textbf{10} & 16 & 11
        & \textbf{10} & 16 & \textbf{10}
        & \textbf{9} & 16 & \textbf{9}
        & \textbf{10} & 16 & \textbf{10}
        \\

        & LS-LN
        & \textbf{16} & 22 & \textbf{16}
        & 16 & 23 & \textbf{15}
        & \textbf{15} & 22 & \textbf{15}
        & 15 & 23 & \textbf{14}
        & 17 & 23 & \textbf{16}
        \\

        & LS-LW
        & \textbf{17} & 25 & \textbf{17}
        & \textbf{17} & 24 & \textbf{17}
        & \textbf{15} & 20 & \textbf{15}
        & \textbf{16} & 25 & 17
        & \textbf{16} & 23 & \textbf{16}
        \\
        \cmidrule(lr){2-17}


        \multirow{3}{*}{\rotatebox[origin=c]{90}{MetaDr.}}
        & Curve
        & \textbf{12} & 16 & 13
        & 13 & 18 & \textbf{12}
        & 14 & 19 & \textbf{13}
        & \textbf{13} & 17 & \textbf{13}
        & 15 & 21 & \textbf{13}
        \\

        & Straight
        & 8 & 10 & \textbf{7}
        & \textbf{7} & 8 & \textbf{7}
        & \textbf{7} & 10 & \textbf{7}
        & \textbf{7} & 9 & \textbf{7}
        & \textbf{7} & 8 & \textbf{7}
        \\

        & Mixed
        & \textbf{9} & 14 & \textbf{9}
        & \textbf{10} & 15 & \textbf{10}
        & 12 & 14 & \textbf{10}
        & 11 & 18 & \textbf{10}
        & 12 & 15 & \textbf{10}
        \\
        \bottomrule

    \end{tabular}

\end{table}

\section{Conclusion}
\label{sec:conclusion}



We discussed the two types of explanations to address the goals of failure analysis: reproduction and avoidance. Previous work introduced sufficient explanations, and we argued that these are well-suited for failure reproduction. We then formulated a new type of explanations---necessary explanations---that is suited for failure avoidance and repair. Both types of explanations can be computed in a single framework of actual causality, giving us a unified approach to failure analysis. We presented an algorithm, and our evaluation included an efficiency estimation and a comparison with the existing methods of counterfactual generation. 

\bibliographystyle{IEEEtran}
\bibliography{references,all}


\clearpage

\begin{appendices}

\section{Evaluated Track Visualization}
\label{app:tracks}

\begin{figure}[h]
\centering
\includegraphics[width=\columnwidth]{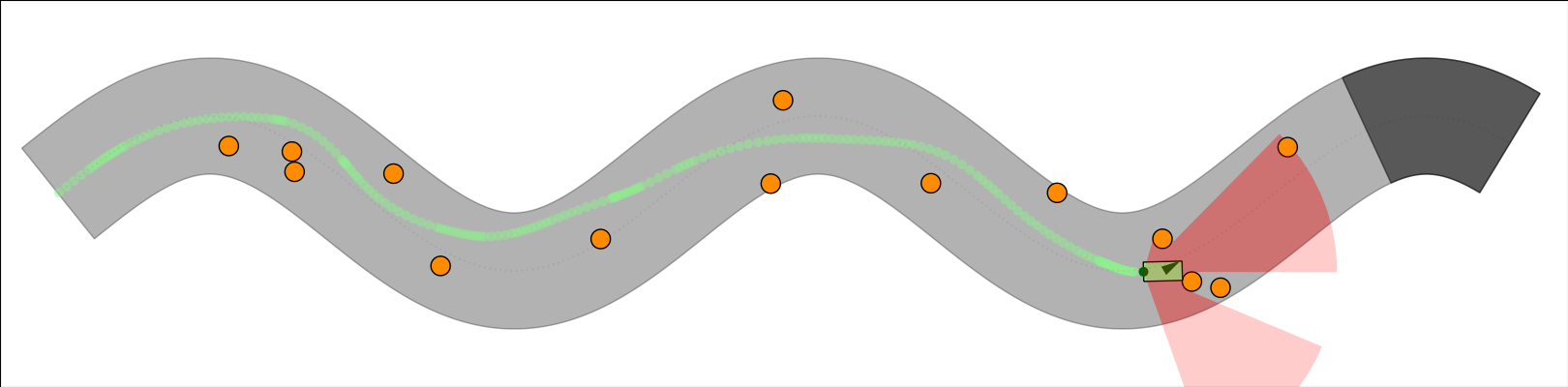}
\caption{Sinusoidal short narrow}
\label{fig:Sin-SN}
\end{figure}

\begin{figure}[h]
\centering
\includegraphics[width=\columnwidth]{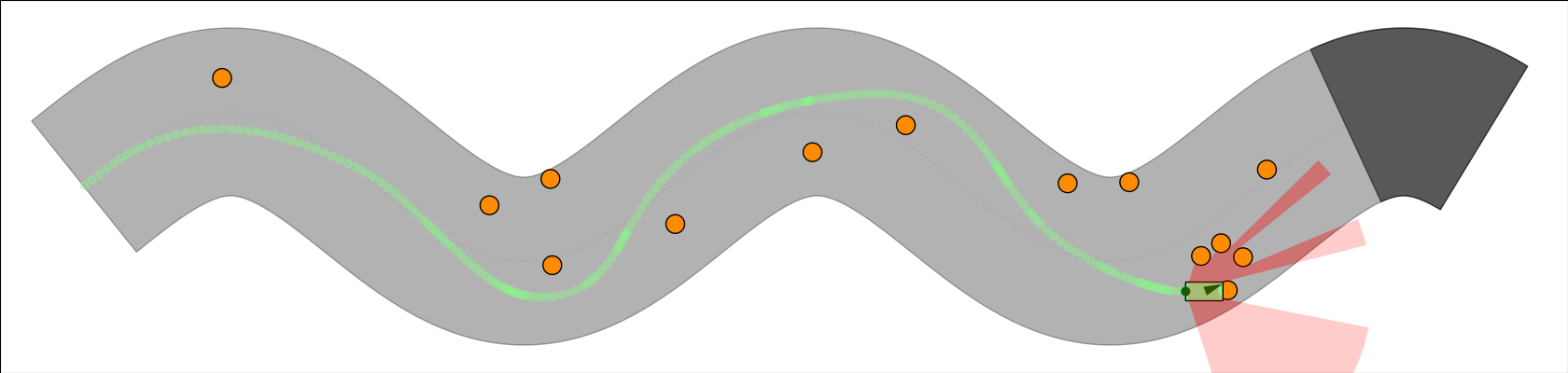}
\caption{Sinusoidal short wide}
\label{fig:Sin-SW}
\end{figure}

\begin{figure}[h]
\centering
\includegraphics[width=\columnwidth]{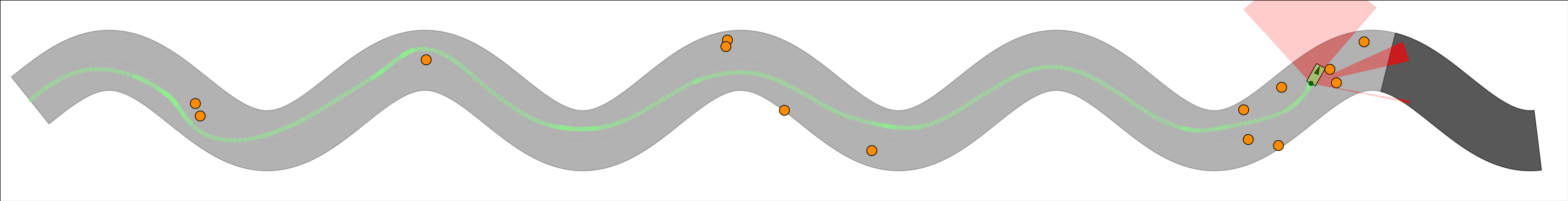}
\caption{Sinusoidal long narrow}
\label{fig:Sin-LN}
\end{figure}

\begin{figure}[h]
\centering
\includegraphics[width=\columnwidth]{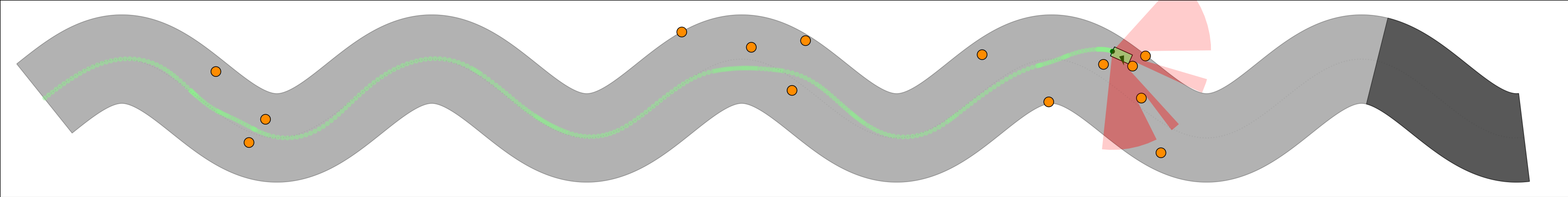}
\caption{Sinusoidal long wide}
\label{fig:Sin-LW}
\end{figure}

\begin{figure}[h]
\centering
\includegraphics[width=\columnwidth]{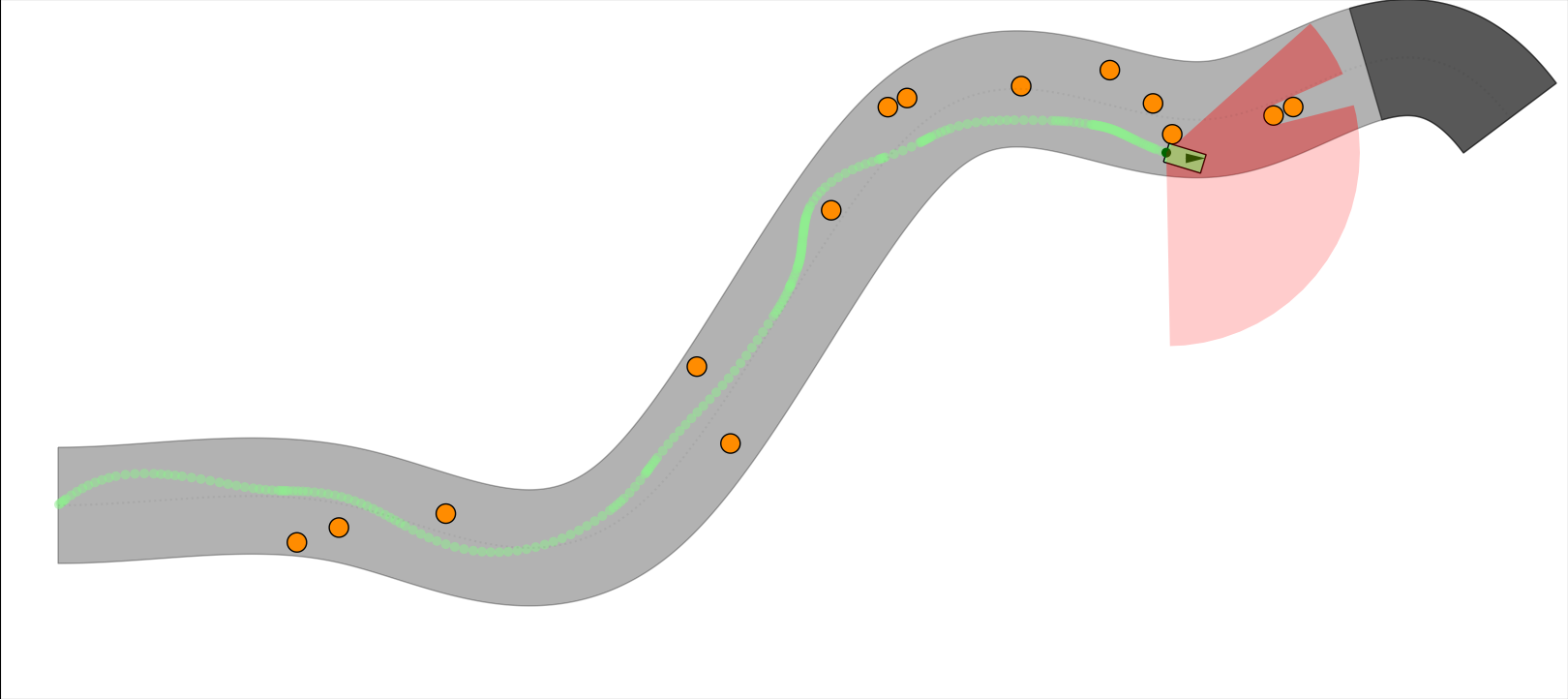}
\caption{Segmented short narrow}
\label{fig:Seg-SN}
\end{figure}

\begin{figure}[h]
\centering
\includegraphics[width=\columnwidth]{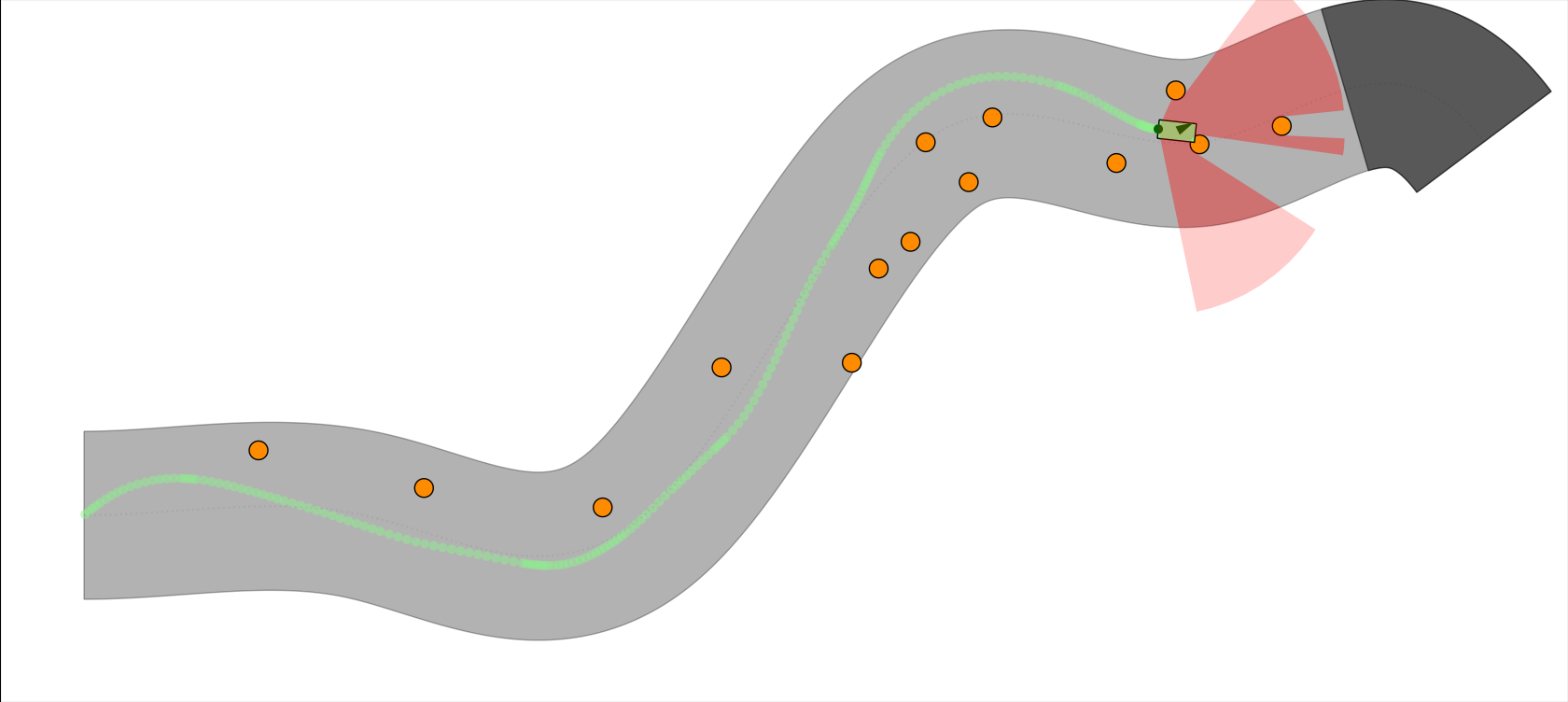}
\caption{Segmented short wide}
\label{fig:Seg-SW}
\end{figure}

\begin{figure}[h]
\centering
\includegraphics[width=\columnwidth]{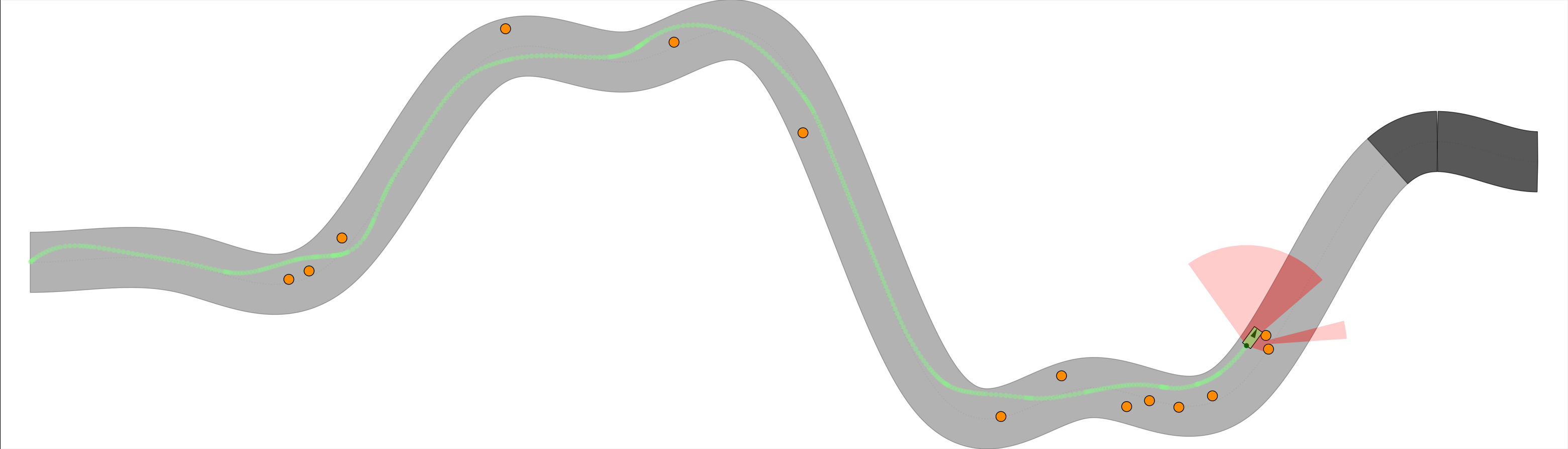}
\caption{Segmented long narrow}
\label{fig:Seg-LN}
\end{figure}

\begin{figure}[h]
\centering
\includegraphics[width=\columnwidth]{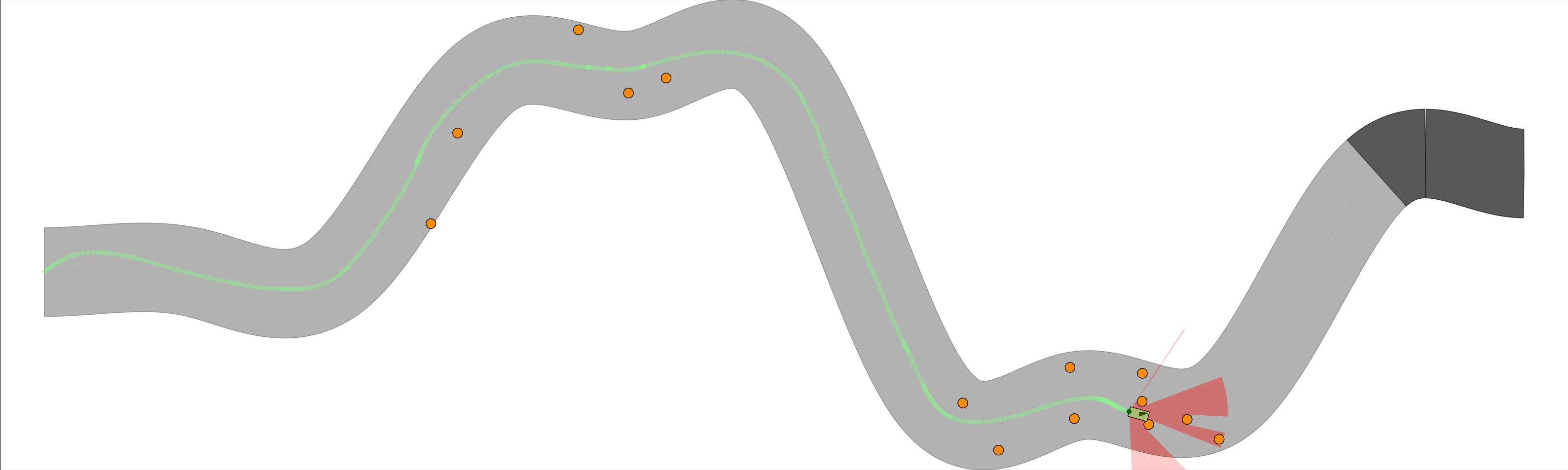}
\caption{Segmented long wide}
\label{fig:Seg-LW}
\end{figure}

\begin{figure}[h]
\centering
\includegraphics[width=\columnwidth]{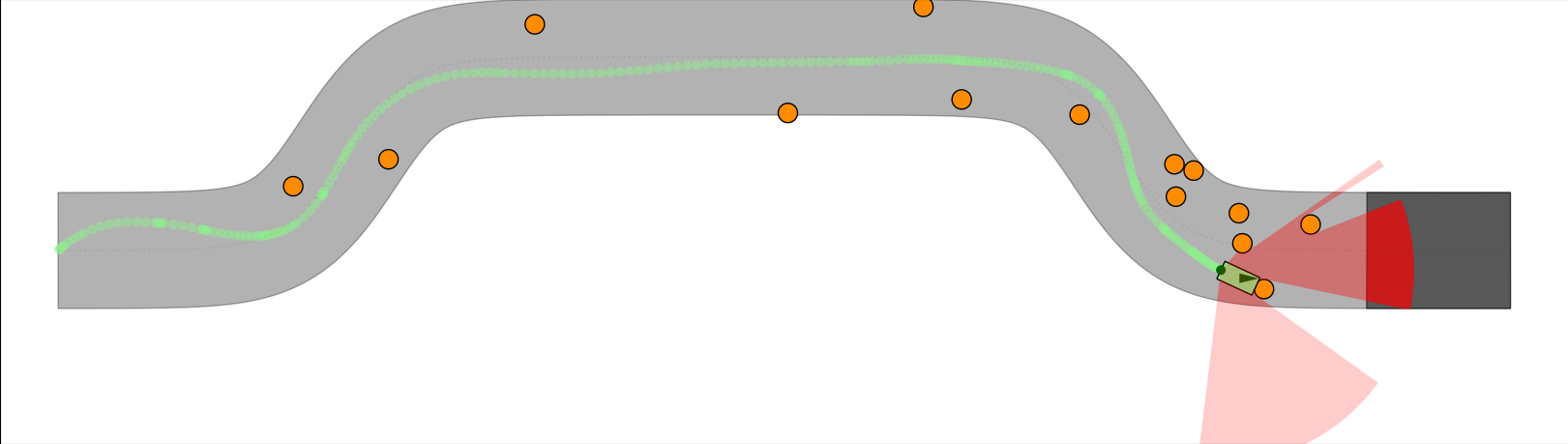}
\caption{Lane Shift short narrow}
\label{fig:LS-SN}
\end{figure}

\begin{figure}[h]
\centering
\includegraphics[width=\columnwidth]{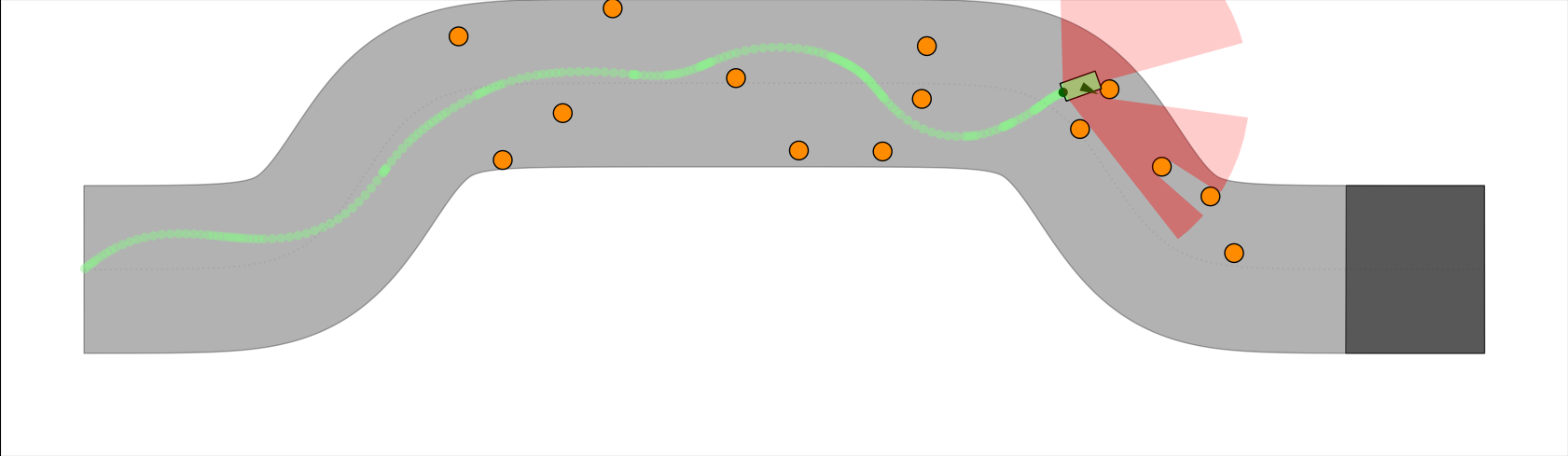}
\caption{Lane Shift short wide}
\label{fig:LS-SW}
\end{figure}

\begin{figure}[h]
\centering
\includegraphics[width=\columnwidth]{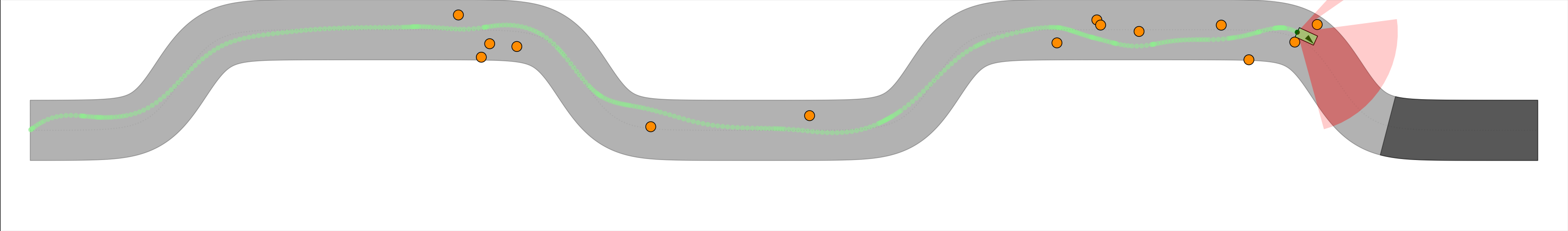}
\caption{Lane Shift long narrow}
\label{fig:LS-LN}
\end{figure}

\begin{figure}[h]
\centering
\includegraphics[width=\columnwidth]{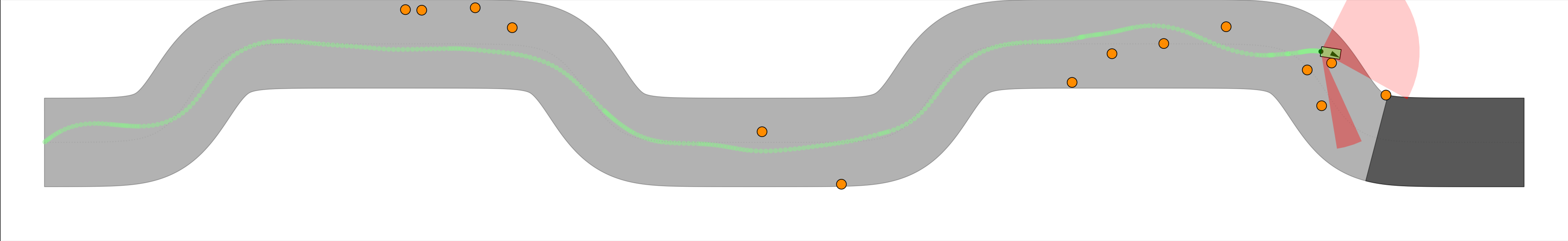}
\caption{Lane Shift long wide}
\label{fig:LS-LW}
\end{figure}

\begin{figure}[h]
\centering
\includegraphics[width=\columnwidth]{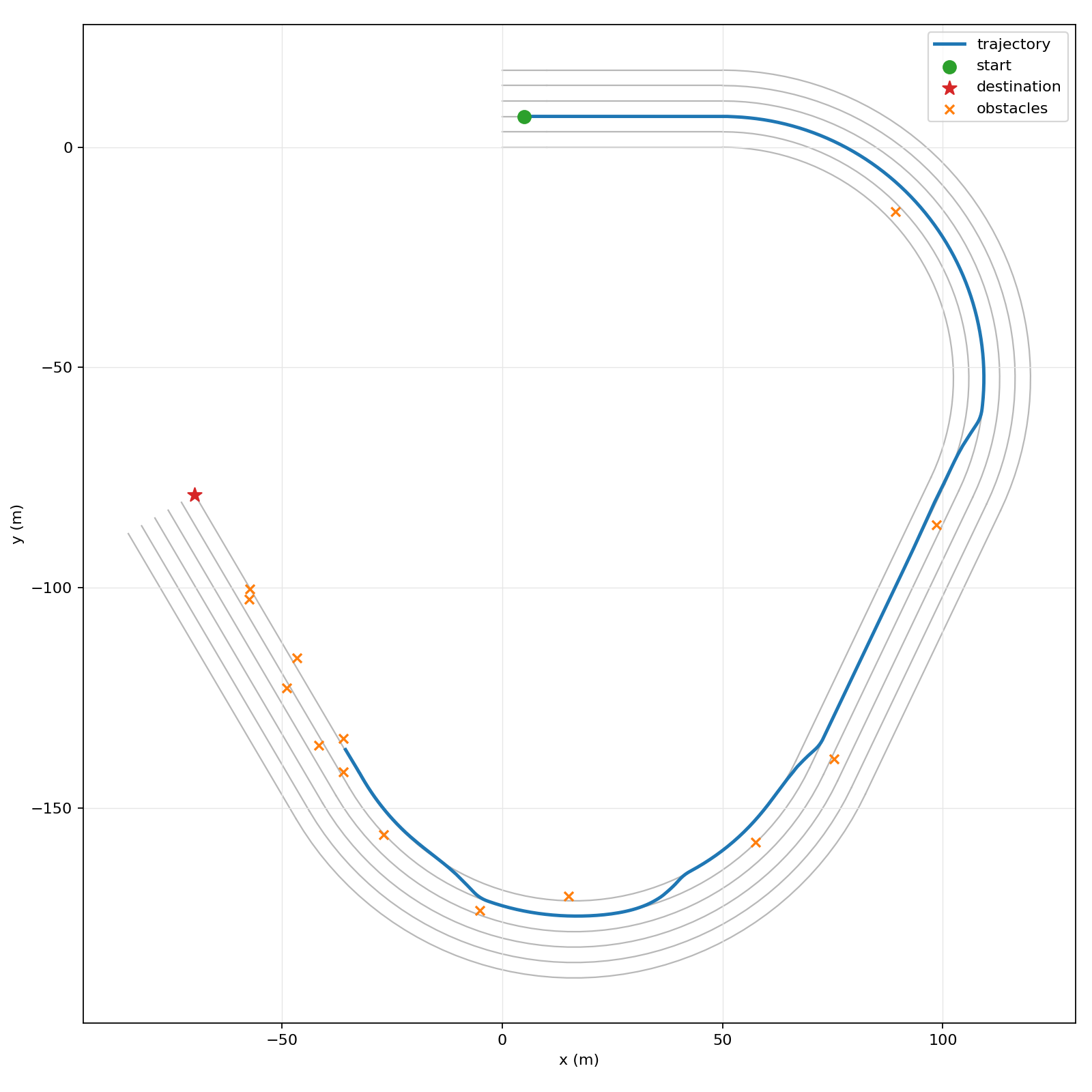}
\caption{MetaDrive Curve}
\label{fig:Curve}
\end{figure}

\begin{figure}[h]
\centering
\includegraphics[width=\columnwidth]{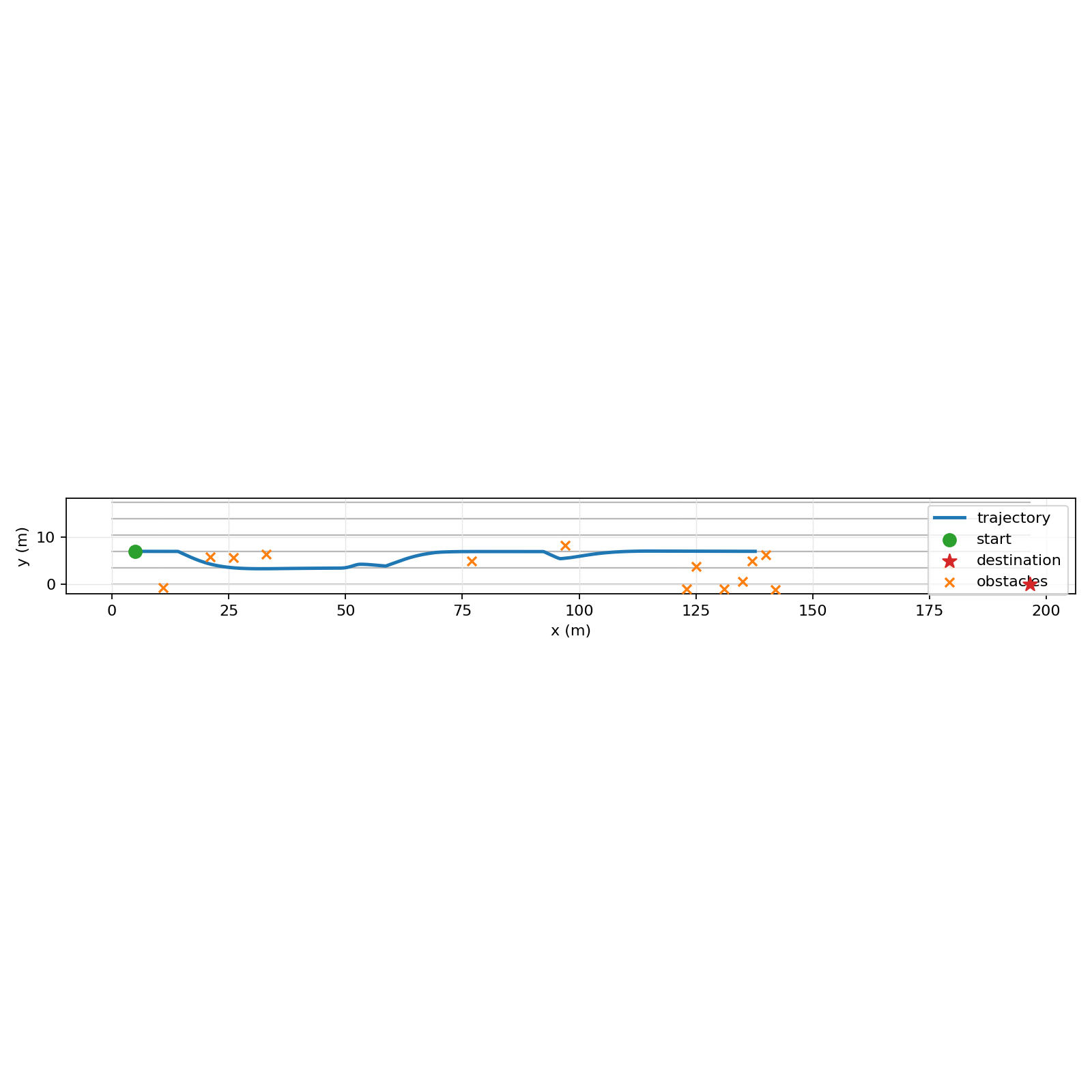}
\caption{MetaDrive Straight}
\label{fig:Straight}
\end{figure}

\begin{figure}[h]
\centering
\includegraphics[width=\columnwidth]{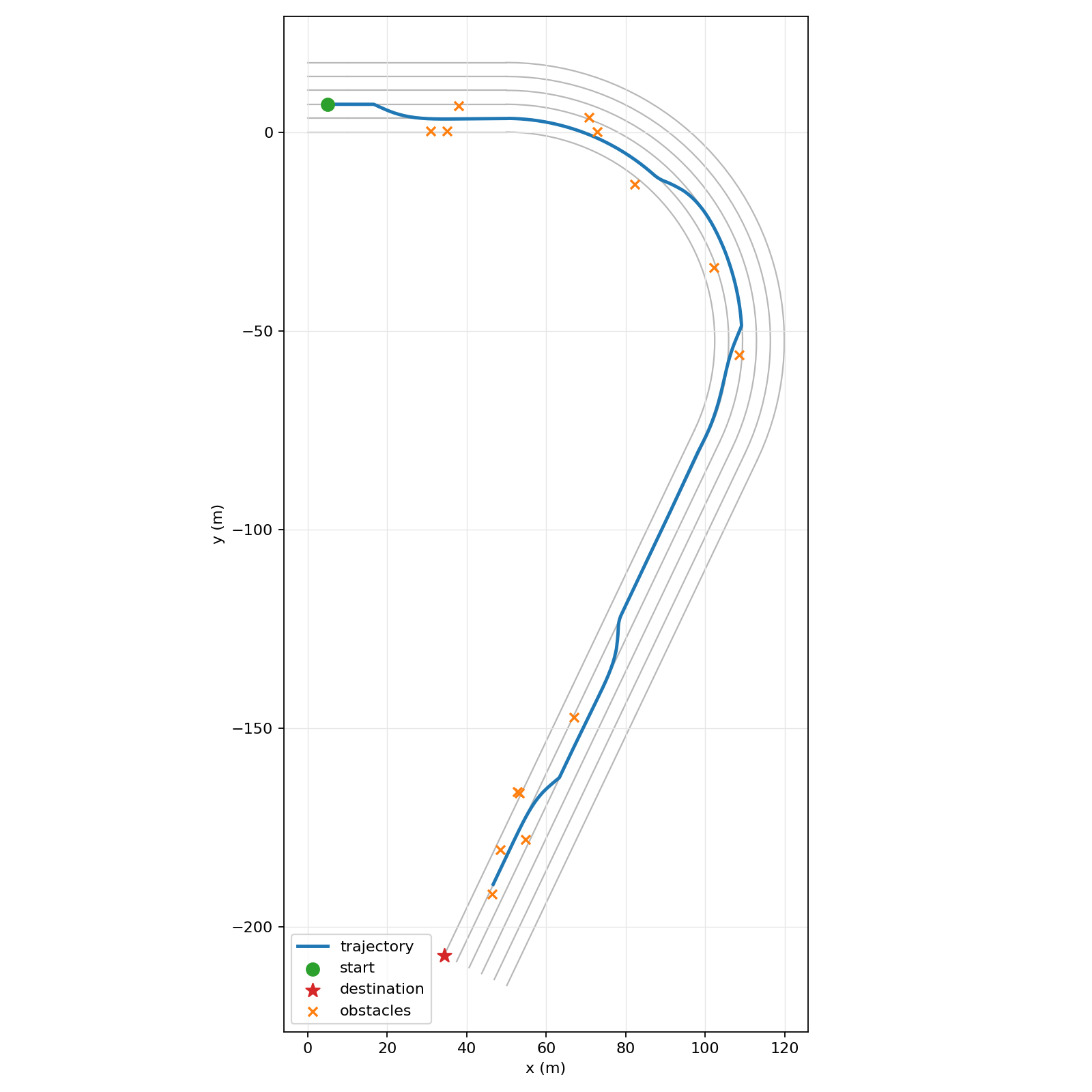}
\caption{MetaDrive Mixed}
\label{fig:Mixed}
\end{figure}

\crefrange{fig:Sin-SN}{fig:Mixed} show all types of evaluated tracks. We note that the visualizations of the MetaDrive environments (\crefrange{fig:Curve}{fig:Mixed}) \textbf{include lanes in both directions}. Hence, the vehicle only drives on the right side of the road.

\clearpage
\section{Execution Time Analysis}
\label{app:time-analysis}

\Cref{fig:execution-time-analysis} shows the execution time in seconds as a function of the number of simulation runs. The measurements were taken for the scripts used in the evaluation presented in \cref{sec:performance-evaluation}, covering the exhaustive and responsibility-guided approaches in the sinusoidal short narrow environment. Importantly, these results demonstrate that the cost of all other computation in the responsibility-guided approach is negligible relative to the number of simulation runs and therefore does not materially affect the overall runtime. Thus, the analysis in \cref{sec:performance-evaluation} focuses on the number of simulation runs. 

\pgfplotsset{
    exhaustiveDots/.style={
        color=red!100,
        only marks,
        mark=*,
        mark size=0.3pt
    },
    exhaustiveLine/.style={
        color=red!100,
        line width=0.2pt,
        no marks
    },
    discreteDots/.style={
        color=blue!100,
        only marks,
        mark=*,
        mark size=0.3pt
    },
    discreteLine/.style={
        color=blue!100,
        line width=0.2pt,
        no marks
    },
    euclideanDots/.style={
        color=blue!100,
        only marks,
        mark=*,
        mark size=0.3pt
    },
    euclideanLine/.style={
        color=blue!100,
        line width=0.4pt,
        no marks
    },
    simulationDots/.style={
        color=green!60!black,
        only marks,
        mark=*,
        mark size=0.3pt
    },
    simulationLine/.style={
        color=green!60!black,
        line width=0.4pt,
        no marks
    }
}

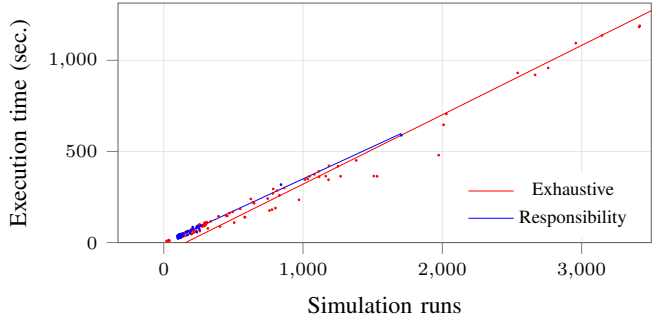
\begin{figure}[h]
    \centering

    \begin{tikzpicture}

        \begin{axis}[
            width=\columnwidth,
            height=0.55\columnwidth,
            ymin=0,
            xmax=3500, 
            xlabel={Simulation runs},
            ylabel={Execution time (sec.)},
            grid=major,
            grid style={black!10},
            axis line style={black!60},
            tick align=outside,
            tick pos=left,
            tick label style={
                font=\scriptsize
            },
            label style={
                font=\small
            },
            legend style={
                font=\scriptsize,
                draw=none,
                at={(0.98,0.02)},
                anchor=south east,
                legend columns=1
            }
        ]


        \addplot[
            exhaustiveDots,
            forget plot
        ] table[
            x=simulation_executions,
            y=execution_time_seconds,
            col sep=comma
        ] {results/necessary_exhaustive_execution_time_analysis.csv};

        \addplot[
            exhaustiveLine,
            forget plot
        ] table[
            x=line_x,
            y=line_y,
            col sep=comma
        ] {results/necessary_exhaustive_execution_time_trend_line.csv};


        \addplot[
            discreteDots,
            forget plot
        ] table[
            x=simulation_executions,
            y=execution_time_seconds,
            col sep=comma
        ] {results/necessary_responsibility_execution_time_analysis.csv};

        \addplot[
            discreteLine,
            forget plot
        ] table[
            x=line_x,
            y=line_y,
            col sep=comma
        ] {results/necessary_responsibility_execution_time_trend_line.csv};








        \addlegendimage{exhaustiveLine}
        \addlegendentry{Exhaustive}

        \addlegendimage{discreteLine}
        \addlegendentry{Responsibility}



        \end{axis}

    \end{tikzpicture}

    \caption{%
        Execution time as a function of the number of simulation runs.
    }
    \label{fig:execution-time-analysis}

\end{figure}

\section{Additional Counterfactual Generation Evaluation Results}
\label{app:counterfactual-results}

\crefrange{tab:average-accuracy-100}{tab:average-accuracy-300} report the counterfactual validity rate under budgets of 100, 200, and 300. \crefrange{fig:counterfactual-distance-15}{fig:counterfactual-distance-18} show the average counterfactual distance as a function of the simulation run budget for environments with 15, 16, 17, and 18 obstacles.

\begin{table}[h]
    \scriptsize
    \centering

    \caption{%
        Counterfactual validity (correctness) rate for the necessary explanations (our approach),
        DeCaF, and random methods with a sample budget of $100$.
    }
    \label{tab:average-accuracy-100}

    \scriptsize
    \setlength{\tabcolsep}{1.6pt}

    \begin{tabular}{c c || ccc ccc ccc ccc ccc}

        \toprule
        \multicolumn{2}{c}{} &
        \multicolumn{3}{c}{\textbf{15}} &
        \multicolumn{3}{c}{\textbf{16}} &
        \multicolumn{3}{c}{\textbf{17}} &
        \multicolumn{3}{c}{\textbf{18}} &
        \multicolumn{3}{c}{\textbf{19}}
        \\
        \cmidrule(lr){3-5}
        \cmidrule(lr){6-8}
        \cmidrule(lr){9-11}
        \cmidrule(lr){12-14}
        \cmidrule(lr){15-17}

        & \textbf{Env.} &
        \textbf{NE} & \textbf{De} & \textbf{Ra} &
        \textbf{NE} & \textbf{De} & \textbf{Ra} &
        \textbf{NE} & \textbf{De} & \textbf{Ra} &
        \textbf{NE} & \textbf{De} & \textbf{Ra} &
        \textbf{NE} & \textbf{De} & \textbf{Ra}
        \\
        \midrule


        \multirow{12}{*}{\rotatebox[origin=c]{90}{LiteRacer}}
        & Sin-SN
        & \textbf{1.0} & .95 & \textbf{1.0}
        & \textbf{1.0} & .95 & \textbf{1.0}
        & \textbf{1.0} & .95 & \textbf{1.0}
        & \textbf{1.0} & .95 & \textbf{1.0}
        & \textbf{1.0} & \textbf{1.0} & \textbf{1.0}
        \\

        & Sin-SW
        & \textbf{1.0} & .80 & \textbf{1.0}
        & \textbf{1.0} & .95 & \textbf{1.0}
        & \textbf{1.0} & .95 & \textbf{1.0}
        & \textbf{1.0} & \textbf{1.0} & \textbf{1.0}
        & \textbf{1.0} & .95 & \textbf{1.0}
        \\

        & Sin-LN
        & \textbf{1.0} & .80 & \textbf{1.0}
        & \textbf{1.0} & .90 & \textbf{1.0}
        & \textbf{1.0} & .95 & \textbf{1.0}
        & \textbf{1.0} & .90 & \textbf{1.0}
        & \textbf{1.0} & .95 & \textbf{1.0}
        \\

        & Sin-LW
        & \textbf{1.0} & .85 & \textbf{1.0}
        & \textbf{1.0} & \textbf{1.0} & \textbf{1.0}
        & \textbf{1.0} & .95 & \textbf{1.0}
        & \textbf{1.0} & .80 & \textbf{1.0}
        & \textbf{1.0} & .80 & \textbf{1.0}
        \\
        \cmidrule(lr){2-17}

        & Seg-SN
        & \textbf{1.0} & .90 & \textbf{1.0}
        & \textbf{1.0} & .95 & \textbf{1.0}
        & \textbf{1.0} & \textbf{1.0} & \textbf{1.0}
        & \textbf{1.0} & .95 & \textbf{1.0}
        & \textbf{1.0} & \textbf{1.0} & \textbf{1.0}
        \\

        & Seg-SW
        & \textbf{1.0} & .95 & \textbf{1.0}
        & \textbf{1.0} & \textbf{1.0} & \textbf{1.0}
        & \textbf{1.0} & .95 & \textbf{1.0}
        & \textbf{1.0} & .90 & \textbf{1.0}
        & \textbf{1.0} & .85 & \textbf{1.0}
        \\

        & Seg-LN
        & \textbf{1.0} & .85 & \textbf{1.0}
        & \textbf{1.0} & .95 & \textbf{1.0}
        & \textbf{1.0} & .85 & \textbf{1.0}
        & \textbf{1.0} & .90 & \textbf{1.0}
        & \textbf{1.0} & .85 & \textbf{1.0}
        \\

        & Seg-LW
        & \textbf{1.0} & .85 & \textbf{1.0}
        & \textbf{1.0} & .90 & \textbf{1.0}
        & \textbf{1.0} & .90 & \textbf{1.0}
        & \textbf{1.0} & .95 & \textbf{1.0}
        & \textbf{1.0} & \textbf{1.0} & \textbf{1.0}
        \\
        \cmidrule(lr){2-17}

        & LS-SN
        & \textbf{1.0} & .90 & \textbf{1.0}
        & \textbf{1.0} & .90 & \textbf{1.0}
        & \textbf{1.0} & .95 & \textbf{1.0}
        & \textbf{1.0} & \textbf{1.0} & \textbf{1.0}
        & \textbf{1.0} & .85 & \textbf{1.0}
        \\

        & LS-SW
        & \textbf{1.0} & .90 & \textbf{1.0}
        & \textbf{1.0} & .95 & \textbf{1.0}
        & \textbf{1.0} & .85 & \textbf{1.0}
        & \textbf{1.0} & .80 & \textbf{1.0}
        & \textbf{1.0} & .95 & \textbf{1.0}
        \\

        & LS-LN
        & \textbf{1.0} & .95 & \textbf{1.0}
        & \textbf{1.0} & \textbf{1.0} & \textbf{1.0}
        & \textbf{1.0} & .90 & \textbf{1.0}
        & \textbf{1.0} & .95 & \textbf{1.0}
        & \textbf{1.0} & .80 & \textbf{1.0}
        \\

        & LS-LW
        & \textbf{1.0} & .90 & \textbf{1.0}
        & \textbf{1.0} & .95 & \textbf{1.0}
        & \textbf{1.0} & .80 & \textbf{1.0}
        & \textbf{1.0} & .80 & \textbf{1.0}
        & \textbf{1.0} & .85 & \textbf{1.0}
        \\
        \cmidrule(lr){2-17}


        \multirow{3}{*}{\rotatebox[origin=c]{90}{MetaDr.}}
        & Curve
        & \textbf{.75} & .45 & .55
        & \textbf{.90} & .55 & .80
        & \textbf{1.0} & .70 & .90
        & \textbf{.80} & .55 & .65
        & \textbf{.90} & .70 & .80
        \\

        & Straight
        & \textbf{.75} & .35 & .40
        & \textbf{.80} & .30 & .35
        & \textbf{.70} & .30 & .40
        & \textbf{.75} & .15 & .45
        & \textbf{.70} & .25 & .30
        \\

        & Mixed
        & \textbf{.65} & .50 & .60
        & \textbf{.75} & .65 & .50
        & \textbf{.90} & .45 & .75
        & \textbf{.90} & .65 & .65
        & \textbf{.80} & .45 & .50
        \\
        \bottomrule

    \end{tabular}

\end{table}

\begin{table}[h]
    \scriptsize
    \centering

    \caption{%
        Counterfactual validity (correctness) rate for the necessary explanations (our approach),
        DeCaF, and random methods with a sample budget of $200$.
    }
    \label{tab:average-accuracy-200}

    \scriptsize
    \setlength{\tabcolsep}{1.6pt}

    \begin{tabular}{c c || ccc ccc ccc ccc ccc}

        \toprule
        \multicolumn{2}{c}{} &
        \multicolumn{3}{c}{\textbf{15}} &
        \multicolumn{3}{c}{\textbf{16}} &
        \multicolumn{3}{c}{\textbf{17}} &
        \multicolumn{3}{c}{\textbf{18}} &
        \multicolumn{3}{c}{\textbf{19}}
        \\
        \cmidrule(lr){3-5}
        \cmidrule(lr){6-8}
        \cmidrule(lr){9-11}
        \cmidrule(lr){12-14}
        \cmidrule(lr){15-17}

        & \textbf{Env.} &
        \textbf{NE} & \textbf{De} & \textbf{Ra} &
        \textbf{NE} & \textbf{De} & \textbf{Ra} &
        \textbf{NE} & \textbf{De} & \textbf{Ra} &
        \textbf{NE} & \textbf{De} & \textbf{Ra} &
        \textbf{NE} & \textbf{De} & \textbf{Ra}
        \\
        \midrule


        \multirow{12}{*}{\rotatebox[origin=c]{90}{LiteRacer}}
        & Sin-SN
        & \textbf{1.0} & .95 & \textbf{1.0}
        & \textbf{1.0} & .95 & \textbf{1.0}
        & \textbf{1.0} & \textbf{1.0} & \textbf{1.0}
        & \textbf{1.0} & .95 & \textbf{1.0}
        & \textbf{1.0} & \textbf{1.0} & \textbf{1.0}
        \\

        & Sin-SW
        & \textbf{1.0} & .90 & \textbf{1.0}
        & \textbf{1.0} & .85 & \textbf{1.0}
        & \textbf{1.0} & .90 & \textbf{1.0}
        & \textbf{1.0} & .95 & \textbf{1.0}
        & \textbf{1.0} & .95 & \textbf{1.0}
        \\

        & Sin-LN
        & \textbf{1.0} & \textbf{1.0} & \textbf{1.0}
        & \textbf{1.0} & .80 & \textbf{1.0}
        & \textbf{1.0} & .85 & \textbf{1.0}
        & \textbf{1.0} & .90 & \textbf{1.0}
        & \textbf{1.0} & .95 & \textbf{1.0}
        \\

        & Sin-LW
        & \textbf{1.0} & .90 & \textbf{1.0}
        & \textbf{1.0} & .95 & \textbf{1.0}
        & \textbf{1.0} & .95 & \textbf{1.0}
        & \textbf{1.0} & .90 & \textbf{1.0}
        & \textbf{1.0} & .95 & \textbf{1.0}
        \\
        \cmidrule(lr){2-17}

        & Seg-SN
        & \textbf{1.0} & .90 & \textbf{1.0}
        & \textbf{1.0} & \textbf{1.0} & \textbf{1.0}
        & \textbf{1.0} & .95 & \textbf{1.0}
        & \textbf{1.0} & .95 & \textbf{1.0}
        & \textbf{1.0} & \textbf{1.0} & \textbf{1.0}
        \\

        & Seg-SW
        & \textbf{1.0} & .85 & \textbf{1.0}
        & \textbf{1.0} & \textbf{1.0} & \textbf{1.0}
        & \textbf{1.0} & \textbf{1.0} & \textbf{1.0}
        & \textbf{1.0} & .95 & \textbf{1.0}
        & \textbf{1.0} & .95 & \textbf{1.0}
        \\

        & Seg-LN
        & \textbf{1.0} & .90 & \textbf{1.0}
        & \textbf{1.0} & .85 & \textbf{1.0}
        & \textbf{1.0} & .90 & \textbf{1.0}
        & \textbf{1.0} & .95 & \textbf{1.0}
        & \textbf{1.0} & .95 & \textbf{1.0}
        \\

        & Seg-LW
        & \textbf{1.0} & .95 & \textbf{1.0}
        & \textbf{1.0} & .75 & \textbf{1.0}
        & \textbf{1.0} & .90 & \textbf{1.0}
        & \textbf{1.0} & .85 & \textbf{1.0}
        & \textbf{1.0} & .95 & \textbf{1.0}
        \\
        \cmidrule(lr){2-17}

        & LS-SN
        & \textbf{1.0} & .85 & \textbf{1.0}
        & \textbf{1.0} & .90 & \textbf{1.0}
        & \textbf{1.0} & .95 & \textbf{1.0}
        & \textbf{1.0} & \textbf{1.0} & \textbf{1.0}
        & \textbf{1.0} & \textbf{1.0} & \textbf{1.0}
        \\

        & LS-SW
        & \textbf{1.0} & \textbf{1.0} & \textbf{1.0}
        & \textbf{1.0} & .90 & \textbf{1.0}
        & \textbf{1.0} & .90 & \textbf{1.0}
        & \textbf{1.0} & \textbf{1.0} & \textbf{1.0}
        & \textbf{1.0} & .90 & \textbf{1.0}
        \\

        & LS-LN
        & \textbf{1.0} & .90 & \textbf{1.0}
        & \textbf{1.0} & .90 & \textbf{1.0}
        & \textbf{1.0} & .95 & \textbf{1.0}
        & \textbf{1.0} & .95 & \textbf{1.0}
        & \textbf{1.0} & .80 & \textbf{1.0}
        \\

        & LS-LW
        & \textbf{1.0} & \textbf{1.0} & \textbf{1.0}
        & \textbf{1.0} & .95 & \textbf{1.0}
        & \textbf{1.0} & .80 & \textbf{1.0}
        & \textbf{1.0} & .95 & \textbf{1.0}
        & \textbf{1.0} & .95 & \textbf{1.0}
        \\
        \cmidrule(lr){2-17}


        \multirow{3}{*}{\rotatebox[origin=c]{90}{MetaDr.}}
        & Curve
        & \textbf{.85} & .50 & .60
        & \textbf{.90} & .75 & .80
        & \textbf{1.0} & .80 & .95
        & \textbf{.85} & .60 & \textbf{.85}
        & \textbf{.85} & .75 & .80
        \\

        & Straight
        & \textbf{.75} & .35 & .45
        & \textbf{.95} & .35 & .40
        & \textbf{.75} & .45 & .50
        & \textbf{.80} & .35 & .50
        & \textbf{.75} & .30 & .35
        \\

        & Mixed
        & \textbf{.70} & .60 & \textbf{.70}
        & \textbf{.80} & .75 & .65
        & \textbf{1.0} & .65 & .95
        & \textbf{.90} & .70 & .80
        & \textbf{.90} & .45 & .65
        \\
        \bottomrule

    \end{tabular}

\end{table}

\begin{table}[h]
    \scriptsize
    \centering

    \caption{%
        Counterfactual validity (correctness) rate for the necessary explanations (our approach),
        DeCaF, and random methods with a sample budget of $300$.
    }
    \label{tab:average-accuracy-300}

    \scriptsize
    \setlength{\tabcolsep}{1.6pt}

    \begin{tabular}{c c || ccc ccc ccc ccc ccc}

        \toprule
        \multicolumn{2}{c}{} &
        \multicolumn{3}{c}{\textbf{15}} &
        \multicolumn{3}{c}{\textbf{16}} &
        \multicolumn{3}{c}{\textbf{17}} &
        \multicolumn{3}{c}{\textbf{18}} &
        \multicolumn{3}{c}{\textbf{19}}
        \\
        \cmidrule(lr){3-5}
        \cmidrule(lr){6-8}
        \cmidrule(lr){9-11}
        \cmidrule(lr){12-14}
        \cmidrule(lr){15-17}

        & \textbf{Env.} &
        \textbf{NE} & \textbf{De} & \textbf{Ra} &
        \textbf{NE} & \textbf{De} & \textbf{Ra} &
        \textbf{NE} & \textbf{De} & \textbf{Ra} &
        \textbf{NE} & \textbf{De} & \textbf{Ra} &
        \textbf{NE} & \textbf{De} & \textbf{Ra}
        \\
        \midrule


        \multirow{12}{*}{\rotatebox[origin=c]{90}{LiteRacer}}
        & Sin-SN
        & \textbf{1.0} & .90 & \textbf{1.0}
        & \textbf{1.0} & \textbf{1.0} & \textbf{1.0}
        & \textbf{1.0} & .95 & \textbf{1.0}
        & \textbf{1.0} & .90 & \textbf{1.0}
        & \textbf{1.0} & \textbf{1.0} & \textbf{1.0}
        \\

        & Sin-SW
        & \textbf{1.0} & \textbf{1.0} & \textbf{1.0}
        & \textbf{1.0} & .95 & \textbf{1.0}
        & \textbf{1.0} & .95 & \textbf{1.0}
        & \textbf{1.0} & .90 & \textbf{1.0}
        & \textbf{1.0} & \textbf{1.0} & \textbf{1.0}
        \\

        & Sin-LN
        & \textbf{1.0} & .95 & \textbf{1.0}
        & \textbf{1.0} & .85 & \textbf{1.0}
        & \textbf{1.0} & \textbf{1.0} & \textbf{1.0}
        & \textbf{1.0} & .95 & \textbf{1.0}
        & \textbf{1.0} & \textbf{1.0} & \textbf{1.0}
        \\

        & Sin-LW
        & \textbf{1.0} & .95 & \textbf{1.0}
        & \textbf{1.0} & .90 & \textbf{1.0}
        & \textbf{1.0} & .95 & \textbf{1.0}
        & \textbf{1.0} & .75 & \textbf{1.0}
        & \textbf{1.0} & .85 & \textbf{1.0}
        \\
        \cmidrule(lr){2-17}

        & Seg-SN
        & \textbf{1.0} & .90 & \textbf{1.0}
        & \textbf{1.0} & .95 & \textbf{1.0}
        & \textbf{1.0} & .95 & \textbf{1.0}
        & \textbf{1.0} & \textbf{1.0} & \textbf{1.0}
        & \textbf{1.0} & .95 & \textbf{1.0}
        \\

        & Seg-SW
        & \textbf{1.0} & .90 & \textbf{1.0}
        & \textbf{1.0} & \textbf{1.0} & \textbf{1.0}
        & \textbf{1.0} & \textbf{1.0} & \textbf{1.0}
        & \textbf{1.0} & .95 & \textbf{1.0}
        & \textbf{1.0} & .95 & \textbf{1.0}
        \\

        & Seg-LN
        & \textbf{1.0} & .90 & \textbf{1.0}
        & \textbf{1.0} & .95 & \textbf{1.0}
        & \textbf{1.0} & .85 & \textbf{1.0}
        & \textbf{1.0} & .90 & \textbf{1.0}
        & \textbf{1.0} & .95 & \textbf{1.0}
        \\

        & Seg-LW
        & \textbf{1.0} & \textbf{1.0} & \textbf{1.0}
        & \textbf{1.0} & .90 & \textbf{1.0}
        & \textbf{1.0} & .90 & \textbf{1.0}
        & \textbf{1.0} & \textbf{1.0} & \textbf{1.0}
        & \textbf{1.0} & .95 & \textbf{1.0}
        \\
        \cmidrule(lr){2-17}

        & LS-SN
        & \textbf{1.0} & .95 & \textbf{1.0}
        & \textbf{1.0} & .85 & \textbf{1.0}
        & \textbf{1.0} & \textbf{1.0} & \textbf{1.0}
        & \textbf{1.0} & .90 & \textbf{1.0}
        & \textbf{1.0} & .95 & \textbf{1.0}
        \\

        & LS-SW
        & \textbf{1.0} & .95 & \textbf{1.0}
        & \textbf{1.0} & .95 & \textbf{1.0}
        & \textbf{1.0} & .85 & \textbf{1.0}
        & \textbf{1.0} & .85 & \textbf{1.0}
        & \textbf{1.0} & .90 & \textbf{1.0}
        \\

        & LS-LN
        & \textbf{1.0} & .90 & \textbf{1.0}
        & \textbf{1.0} & .90 & \textbf{1.0}
        & \textbf{1.0} & \textbf{1.0} & \textbf{1.0}
        & \textbf{1.0} & .95 & \textbf{1.0}
        & \textbf{1.0} & .95 & \textbf{1.0}
        \\

        & LS-LW
        & \textbf{1.0} & .85 & \textbf{1.0}
        & \textbf{1.0} & .90 & \textbf{1.0}
        & \textbf{1.0} & .85 & \textbf{1.0}
        & \textbf{1.0} & .90 & \textbf{1.0}
        & \textbf{1.0} & .90 & \textbf{1.0}
        \\
        \cmidrule(lr){2-17}


        \multirow{3}{*}{\rotatebox[origin=c]{90}{MetaDr.}}
        & Curve
        & \textbf{.85} & .55 & .65
        & \textbf{.90} & .80 & .85
        & \textbf{1.0} & .75 & .95
        & .85 & .65 & \textbf{.90}
        & \textbf{1.0} & .75 & .90
        \\

        & Straight
        & \textbf{.75} & .40 & .45
        & \textbf{.95} & .35 & .50
        & \textbf{.75} & .60 & .55
        & \textbf{.75} & .35 & .55
        & \textbf{.75} & .35 & .40
        \\

        & Mixed
        & \textbf{.75} & .65 & .70
        & \textbf{.85} & .75 & .75
        & \textbf{1.0} & .75 & .95
        & \textbf{.95} & .70 & .85
        & \textbf{.95} & .45 & .75
        \\
        \bottomrule

    \end{tabular}

\end{table}

\pgfplotsset{
    necessaryDistance/.style={
        color=blue!80,
        very thick,
        mark=*,
        mark size=1.3pt
    },
    decafDistance/.style={
        color=red!80,
        thick,
        mark=square*,
        mark size=1.3pt
    },
    randomDistance/.style={
        color=black!70,
        thick,
        mark=triangle*,
        mark size=1.5pt
    }
}

\begin{figure*}[t]
    \centering

    \begin{tikzpicture}

        \begin{groupplot}[
            group style={
                group name=plots,
                group size=6 by 3,
                horizontal sep=7.5mm,
                vertical sep=3mm,
            },
            width=0.218\textwidth,
            height=0.185\textwidth,
            xmin=80,
            xmax=420,
            xtick={100,200,300,400},
            scaled y ticks=false,
            tick align=outside,
            tick pos=left,
            grid=major,
            grid style={black!10},
            axis line style={black!60},
            tick label style={font=\scriptsize},
            label style={font=\small},
            title style={font=\small},
        ]


        \nextgroupplot[
            title={Short Narrow},
            ylabel={Distance},
            xticklabels=\empty,
            legend columns=5,
            legend to name=commonlegend15,
            legend style={
                font=\small,
                draw=none,
                /tikz/every even column/.append style={
                    column sep=3mm
                }
            }
        ]
        \plotdistancewithlegend{results/sinusoidal_short_narrow_average_distance_obs_15.csv}

        \nextgroupplot[
            title={Short Wide},
            xticklabels=\empty,
        ]
        \plotdistance{results/sinusoidal_short_wide_average_distance_obs_15.csv}

        \nextgroupplot[
            title={Long Narrow},
            xticklabels=\empty,
        ]
        \plotdistance{results/sinusoidal_long_narrow_average_distance_obs_15.csv}

        \nextgroupplot[
            title={Long Wide},
            xticklabels=\empty,
        ]
        \plotdistance{results/sinusoidal_long_wide_average_distance_obs_15.csv}

        \nextgroupplot[
            hide axis,
            width=0.11\textwidth
        ]

        \nextgroupplot[
            ylabel={Distance},
            xticklabels=\empty
        ]
        \plotdistance{results/metadrive_curve_average_distance_obs_15.csv}


        \nextgroupplot[
            ylabel={Distance},
            xticklabels=\empty
        ]
        \plotdistance{results/segmented_short_narrow_average_distance_obs_15.csv}

        \nextgroupplot[
            xticklabels=\empty,
        ]
        \plotdistance{results/segmented_short_wide_average_distance_obs_15.csv}

        \nextgroupplot[
            xticklabels=\empty,
        ]
        \plotdistance{results/segmented_long_narrow_average_distance_obs_15.csv}

        \nextgroupplot[
            xticklabels=\empty,
        ]
        \plotdistance{results/segmented_long_wide_average_distance_obs_15.csv}

        \nextgroupplot[
            hide axis,
            width=0.11\textwidth
        ]

        \nextgroupplot[
            ylabel={Distance},
            xticklabels=\empty
        ]
        \plotdistance{results/metadrive_straight_average_distance_obs_15.csv}


        \nextgroupplot[
            xlabel={Number of Samples},
            ylabel={Distance}
        ]
        \plotdistance{results/lane_shift_short_narrow_average_distance_obs_15.csv}

        \nextgroupplot[
            xlabel={Number of Samples},
        ]
        \plotdistance{results/lane_shift_short_wide_average_distance_obs_15.csv}

        \nextgroupplot[
            xlabel={Number of Samples},
        ]
        \plotdistance{results/lane_shift_long_narrow_average_distance_obs_15.csv}

        \nextgroupplot[
            xlabel={Number of Samples},
        ]
        \plotdistance{results/lane_shift_long_wide_average_distance_obs_15.csv}

        \nextgroupplot[
            hide axis,
            width=0.11\textwidth
        ]

        \nextgroupplot[
            xlabel={Number of Samples},
            ylabel={Distance},
            xtick={100,200,300,400},
            xticklabels={100,200,300,400}
        ]
        \plotdistance{results/metadrive_mixed_average_distance_obs_15.csv}

        \end{groupplot}


        \node[
            anchor=south,
            font=\small
        ] at ([yshift=7mm]
             $(plots c1r1.north)!0.5!(plots c4r1.north)$)
        {LiteRacer};

        \node[
            anchor=south,
            font=\small
        ] at ([yshift=7mm]
             $(plots c6r1.north)$)
        {MetaDrive};


        \node[
            rotate=90,
            anchor=south,
            font=\small
        ] at ([xshift=-12mm]$(plots c1r1.west)$)
        {Sinusoidal};

        \node[
            rotate=90,
            anchor=south,
            font=\small
        ] at ([xshift=-12mm]$(plots c1r2.west)$)
        {Segmented};

        \node[
            rotate=90,
            anchor=south,
            font=\small
        ] at ([xshift=-12mm]$(plots c1r3.west)$)
        {Lane Shift};


        \node[
            rotate=90,
            anchor=south,
            font=\small
        ] at ([xshift=-12mm]$(plots c6r1.west)$)
        {Curve};

        \node[
            rotate=90,
            anchor=south,
            font=\small
        ] at ([xshift=-12mm]$(plots c6r2.west)$)
        {Straight};

        \node[
            rotate=90,
            anchor=south,
            font=\small
        ] at ([xshift=-12mm]$(plots c6r3.west)$)
        {Mixed};


        \node[
            anchor=north
        ] at ([yshift=-10mm]
             $(plots c1r3.south)!0.5!(plots c6r3.south)$)
        {\pgfplotslegendfromname{commonlegend15}};

    \end{tikzpicture}

    \caption{%
        Quality of counterfactuals as a function of the number of simulation run samples for settings with $15$ obstacles, for necessary explanations (our approach), DeCaF, and random methods, measured by the average distance from the original example (lower is better).
    }
    \label{fig:counterfactual-distance-15}

\end{figure*}
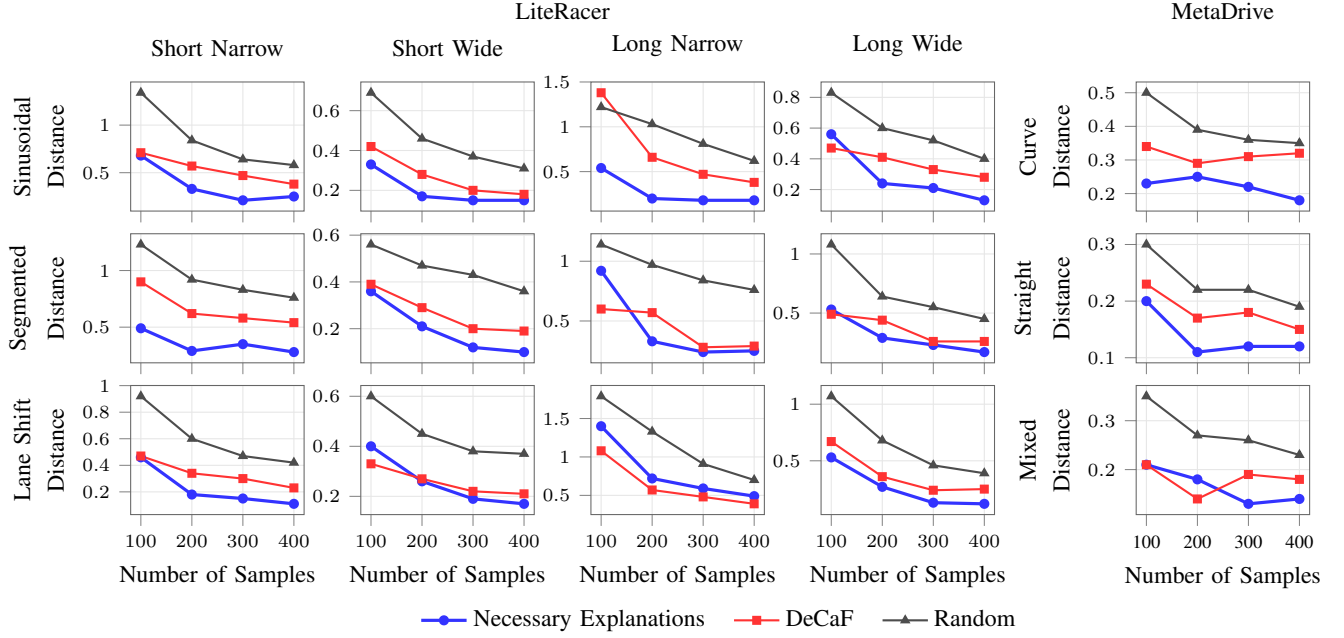

\pgfplotsset{
    necessaryDistance/.style={
        color=blue!80,
        very thick,
        mark=*,
        mark size=1.3pt
    },
    decafDistance/.style={
        color=red!80,
        thick,
        mark=square*,
        mark size=1.3pt
    },
    randomDistance/.style={
        color=black!70,
        thick,
        mark=triangle*,
        mark size=1.5pt
    }
}

\begin{figure*}[t]
    \centering

    \begin{tikzpicture}

        \begin{groupplot}[
            group style={
                group name=plots,
                group size=6 by 3,
                horizontal sep=7.5mm,
                vertical sep=3mm,
            },
            width=0.218\textwidth,
            height=0.185\textwidth,
            xmin=80,
            xmax=420,
            xtick={100,200,300,400},
            scaled y ticks=false,
            tick align=outside,
            tick pos=left,
            grid=major,
            grid style={black!10},
            axis line style={black!60},
            tick label style={font=\scriptsize},
            label style={font=\small},
            title style={font=\small},
        ]


        \nextgroupplot[
            title={Short Narrow},
            ylabel={Distance},
            xticklabels=\empty,
            legend columns=5,
            legend to name=commonlegend16,
            legend style={
                font=\small,
                draw=none,
                /tikz/every even column/.append style={
                    column sep=3mm
                }
            }
        ]
        \plotdistancewithlegend{results/sinusoidal_short_narrow_average_distance_obs_16.csv}

        \nextgroupplot[
            title={Short Wide},
            xticklabels=\empty,
        ]
        \plotdistance{results/sinusoidal_short_wide_average_distance_obs_16.csv}

        \nextgroupplot[
            title={Long Narrow},
            xticklabels=\empty,
        ]
        \plotdistance{results/sinusoidal_long_narrow_average_distance_obs_16.csv}

        \nextgroupplot[
            title={Long Wide},
            xticklabels=\empty,
        ]
        \plotdistance{results/sinusoidal_long_wide_average_distance_obs_16.csv}

        \nextgroupplot[
            hide axis,
            width=0.11\textwidth
        ]

        \nextgroupplot[
            ylabel={Distance},
            xticklabels=\empty
        ]
        \plotdistance{results/metadrive_curve_average_distance_obs_16.csv}


        \nextgroupplot[
            ylabel={Distance},
            xticklabels=\empty
        ]
        \plotdistance{results/segmented_short_narrow_average_distance_obs_16.csv}

        \nextgroupplot[
            xticklabels=\empty,
        ]
        \plotdistance{results/segmented_short_wide_average_distance_obs_16.csv}

        \nextgroupplot[
            xticklabels=\empty,
        ]
        \plotdistance{results/segmented_long_narrow_average_distance_obs_16.csv}

        \nextgroupplot[
            xticklabels=\empty,
        ]
        \plotdistance{results/segmented_long_wide_average_distance_obs_16.csv}

        \nextgroupplot[
            hide axis,
            width=0.11\textwidth
        ]

        \nextgroupplot[
            ylabel={Distance},
            xticklabels=\empty
        ]
        \plotdistance{results/metadrive_straight_average_distance_obs_16.csv}


        \nextgroupplot[
            xlabel={Number of Samples},
            ylabel={Distance}
        ]
        \plotdistance{results/lane_shift_short_narrow_average_distance_obs_16.csv}

        \nextgroupplot[
            xlabel={Number of Samples},
        ]
        \plotdistance{results/lane_shift_short_wide_average_distance_obs_16.csv}

        \nextgroupplot[
            xlabel={Number of Samples},
        ]
        \plotdistance{results/lane_shift_long_narrow_average_distance_obs_16.csv}

        \nextgroupplot[
            xlabel={Number of Samples},
        ]
        \plotdistance{results/lane_shift_long_wide_average_distance_obs_16.csv}

        \nextgroupplot[
            hide axis,
            width=0.11\textwidth
        ]

        \nextgroupplot[
            xlabel={Number of Samples},
            ylabel={Distance},
            xtick={100,200,300,400},
            xticklabels={100,200,300,400}
        ]
        \plotdistance{results/metadrive_mixed_average_distance_obs_16.csv}

        \end{groupplot}


        \node[
            anchor=south,
            font=\small
        ] at ([yshift=7mm]
             $(plots c1r1.north)!0.5!(plots c4r1.north)$)
        {LiteRacer};

        \node[
            anchor=south,
            font=\small
        ] at ([yshift=7mm]
             $(plots c6r1.north)$)
        {MetaDrive};


        \node[
            rotate=90,
            anchor=south,
            font=\small
        ] at ([xshift=-12mm]$(plots c1r1.west)$)
        {Sinusoidal};

        \node[
            rotate=90,
            anchor=south,
            font=\small
        ] at ([xshift=-12mm]$(plots c1r2.west)$)
        {Segmented};

        \node[
            rotate=90,
            anchor=south,
            font=\small
        ] at ([xshift=-12mm]$(plots c1r3.west)$)
        {Lane Shift};


        \node[
            rotate=90,
            anchor=south,
            font=\small
        ] at ([xshift=-12mm]$(plots c6r1.west)$)
        {Curve};

        \node[
            rotate=90,
            anchor=south,
            font=\small
        ] at ([xshift=-12mm]$(plots c6r2.west)$)
        {Straight};

        \node[
            rotate=90,
            anchor=south,
            font=\small
        ] at ([xshift=-12mm]$(plots c6r3.west)$)
        {Mixed};


        \node[
            anchor=north
        ] at ([yshift=-10mm]
             $(plots c1r3.south)!0.5!(plots c6r3.south)$)
        {\pgfplotslegendfromname{commonlegend16}};

    \end{tikzpicture}

    \caption{%
        Quality of counterfactuals as a function of the number of simulation run samples for settings with $16$ obstacles, for necessary explanations (our approach), DeCaF, and random methods, measured by the average distance from the original example (lower is better).
    }
    \label{fig:counterfactual-distance-16}

\end{figure*}
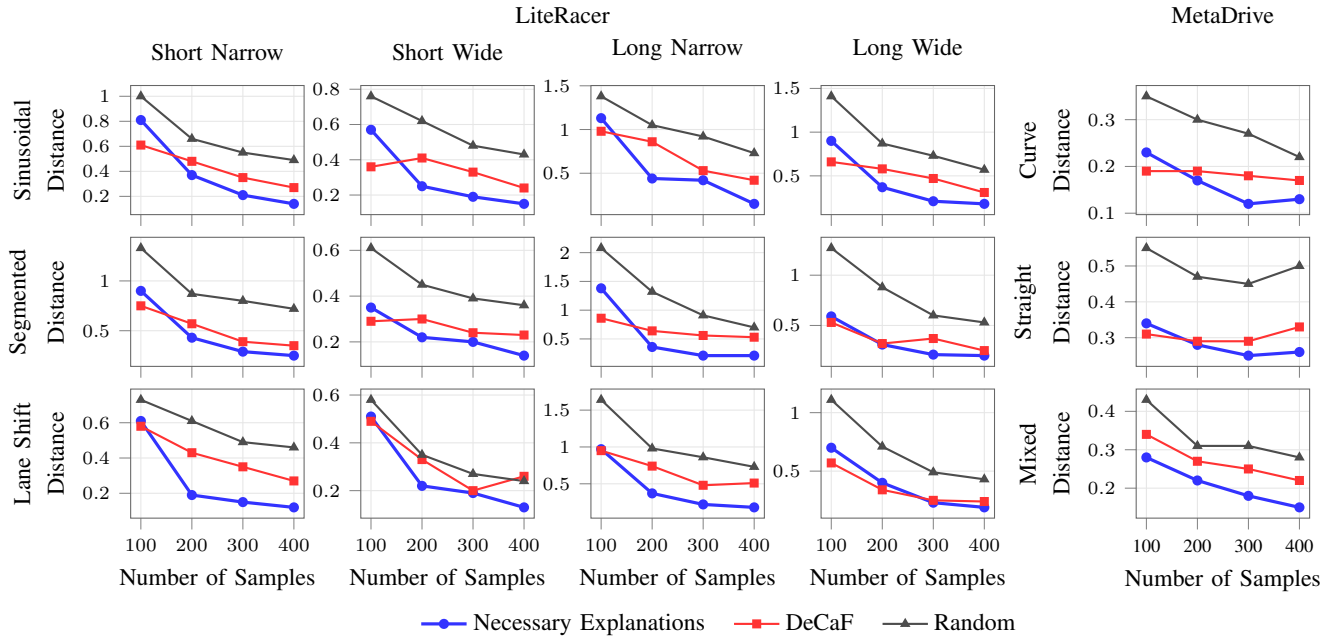

\pgfplotsset{
    necessaryDistance/.style={
        color=blue!80,
        very thick,
        mark=*,
        mark size=1.3pt
    },
    decafDistance/.style={
        color=red!80,
        thick,
        mark=square*,
        mark size=1.3pt
    },
    randomDistance/.style={
        color=black!70,
        thick,
        mark=triangle*,
        mark size=1.5pt
    }
}

\begin{figure*}[t]
    \centering

    \begin{tikzpicture}

        \begin{groupplot}[
            group style={
                group name=plots,
                group size=6 by 3,
                horizontal sep=7.5mm,
                vertical sep=3mm,
            },
            width=0.218\textwidth,
            height=0.185\textwidth,
            xmin=80,
            xmax=420,
            xtick={100,200,300,400},
            scaled y ticks=false,
            tick align=outside,
            tick pos=left,
            grid=major,
            grid style={black!10},
            axis line style={black!60},
            tick label style={font=\scriptsize},
            label style={font=\small},
            title style={font=\small},
        ]


        \nextgroupplot[
            title={Short Narrow},
            ylabel={Distance},
            xticklabels=\empty,
            legend columns=5,
            legend to name=commonlegend17,
            legend style={
                font=\small,
                draw=none,
                /tikz/every even column/.append style={
                    column sep=3mm
                }
            }
        ]
        \plotdistancewithlegend{results/sinusoidal_short_narrow_average_distance_obs_17.csv}

        \nextgroupplot[
            title={Short Wide},
            xticklabels=\empty,
        ]
        \plotdistance{results/sinusoidal_short_wide_average_distance_obs_17.csv}

        \nextgroupplot[
            title={Long Narrow},
            xticklabels=\empty,
        ]
        \plotdistance{results/sinusoidal_long_narrow_average_distance_obs_17.csv}

        \nextgroupplot[
            title={Long Wide},
            xticklabels=\empty,
        ]
        \plotdistance{results/sinusoidal_long_wide_average_distance_obs_17.csv}

        \nextgroupplot[
            hide axis,
            width=0.11\textwidth
        ]

        \nextgroupplot[
            ylabel={Distance},
            xticklabels=\empty
        ]
        \plotdistance{results/metadrive_curve_average_distance_obs_17.csv}


        \nextgroupplot[
            ylabel={Distance},
            xticklabels=\empty
        ]
        \plotdistance{results/segmented_short_narrow_average_distance_obs_17.csv}

        \nextgroupplot[
            xticklabels=\empty,
        ]
        \plotdistance{results/segmented_short_wide_average_distance_obs_17.csv}

        \nextgroupplot[
            xticklabels=\empty,
        ]
        \plotdistance{results/segmented_long_narrow_average_distance_obs_17.csv}

        \nextgroupplot[
            xticklabels=\empty,
        ]
        \plotdistance{results/segmented_long_wide_average_distance_obs_17.csv}

        \nextgroupplot[
            hide axis,
            width=0.11\textwidth
        ]

        \nextgroupplot[
            ylabel={Distance},
            xticklabels=\empty
        ]
        \plotdistance{results/metadrive_straight_average_distance_obs_17.csv}


        \nextgroupplot[
            xlabel={Number of Samples},
            ylabel={Distance}
        ]
        \plotdistance{results/lane_shift_short_narrow_average_distance_obs_17.csv}

        \nextgroupplot[
            xlabel={Number of Samples},
        ]
        \plotdistance{results/lane_shift_short_wide_average_distance_obs_17.csv}

        \nextgroupplot[
            xlabel={Number of Samples},
        ]
        \plotdistance{results/lane_shift_long_narrow_average_distance_obs_17.csv}

        \nextgroupplot[
            xlabel={Number of Samples},
        ]
        \plotdistance{results/lane_shift_long_wide_average_distance_obs_17.csv}

        \nextgroupplot[
            hide axis,
            width=0.11\textwidth
        ]

        \nextgroupplot[
            xlabel={Number of Samples},
            ylabel={Distance},
            xtick={100,200,300,400},
            xticklabels={100,200,300,400}
        ]
        \plotdistance{results/metadrive_mixed_average_distance_obs_17.csv}

        \end{groupplot}


        \node[
            anchor=south,
            font=\small
        ] at ([yshift=7mm]
             $(plots c1r1.north)!0.5!(plots c4r1.north)$)
        {LiteRacer};

        \node[
            anchor=south,
            font=\small
        ] at ([yshift=7mm]
             $(plots c6r1.north)$)
        {MetaDrive};


        \node[
            rotate=90,
            anchor=south,
            font=\small
        ] at ([xshift=-12mm]$(plots c1r1.west)$)
        {Sinusoidal};

        \node[
            rotate=90,
            anchor=south,
            font=\small
        ] at ([xshift=-12mm]$(plots c1r2.west)$)
        {Segmented};

        \node[
            rotate=90,
            anchor=south,
            font=\small
        ] at ([xshift=-12mm]$(plots c1r3.west)$)
        {Lane Shift};


        \node[
            rotate=90,
            anchor=south,
            font=\small
        ] at ([xshift=-12mm]$(plots c6r1.west)$)
        {Curve};

        \node[
            rotate=90,
            anchor=south,
            font=\small
        ] at ([xshift=-12mm]$(plots c6r2.west)$)
        {Straight};

        \node[
            rotate=90,
            anchor=south,
            font=\small
        ] at ([xshift=-12mm]$(plots c6r3.west)$)
        {Mixed};


        \node[
            anchor=north
        ] at ([yshift=-10mm]
             $(plots c1r3.south)!0.5!(plots c6r3.south)$)
        {\pgfplotslegendfromname{commonlegend17}};

    \end{tikzpicture}

    \caption{%
        Quality of counterfactuals as a function of the number of simulation run samples for settings with $17$ obstacles, for necessary explanations (our approach), DeCaF, and random methods, measured by the average distance from the original example (lower is better).
    }
    \label{fig:counterfactual-distance-17}

\end{figure*}
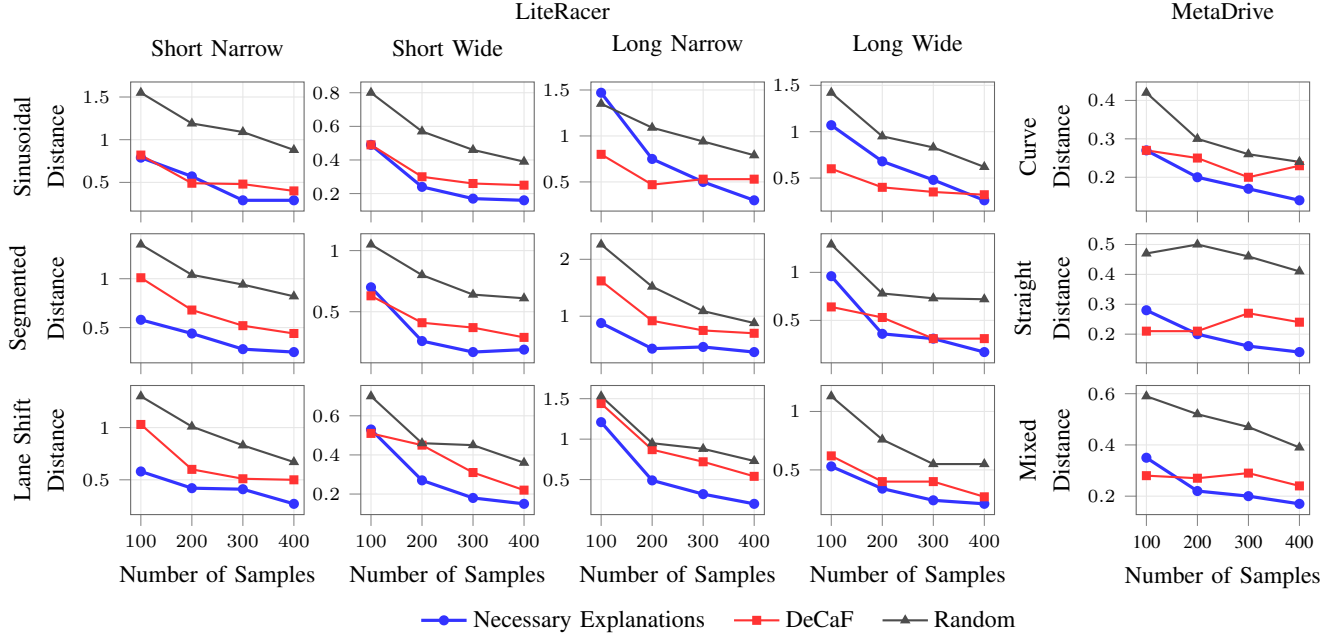

\pgfplotsset{
    necessaryDistance/.style={
        color=blue!80,
        very thick,
        mark=*,
        mark size=1.3pt
    },
    decafDistance/.style={
        color=red!80,
        thick,
        mark=square*,
        mark size=1.3pt
    },
    randomDistance/.style={
        color=black!70,
        thick,
        mark=triangle*,
        mark size=1.5pt
    }
}

\begin{figure*}[t]
    \centering

    \begin{tikzpicture}

        \begin{groupplot}[
            group style={
                group name=plots,
                group size=6 by 3,
                horizontal sep=7.5mm,
                vertical sep=3mm,
            },
            width=0.218\textwidth,
            height=0.185\textwidth,
            xmin=80,
            xmax=420,
            xtick={100,200,300,400},
            scaled y ticks=false,
            tick align=outside,
            tick pos=left,
            grid=major,
            grid style={black!10},
            axis line style={black!60},
            tick label style={font=\scriptsize},
            label style={font=\small},
            title style={font=\small},
        ]


        \nextgroupplot[
            title={Short Narrow},
            ylabel={Distance},
            xticklabels=\empty,
            legend columns=5,
            legend to name=commonlegend18,
            legend style={
                font=\small,
                draw=none,
                /tikz/every even column/.append style={
                    column sep=3mm
                }
            }
        ]
        \plotdistancewithlegend{results/sinusoidal_short_narrow_average_distance_obs_18.csv}

        \nextgroupplot[
            title={Short Wide},
            xticklabels=\empty,
        ]
        \plotdistance{results/sinusoidal_short_wide_average_distance_obs_18.csv}

        \nextgroupplot[
            title={Long Narrow},
            xticklabels=\empty,
        ]
        \plotdistance{results/sinusoidal_long_narrow_average_distance_obs_18.csv}

        \nextgroupplot[
            title={Long Wide},
            xticklabels=\empty,
        ]
        \plotdistance{results/sinusoidal_long_wide_average_distance_obs_18.csv}

        \nextgroupplot[
            hide axis,
            width=0.11\textwidth
        ]

        \nextgroupplot[
            ylabel={Distance},
            xticklabels=\empty
        ]
        \plotdistance{results/metadrive_curve_average_distance_obs_18.csv}


        \nextgroupplot[
            ylabel={Distance},
            xticklabels=\empty
        ]
        \plotdistance{results/segmented_short_narrow_average_distance_obs_18.csv}

        \nextgroupplot[
            xticklabels=\empty,
        ]
        \plotdistance{results/segmented_short_wide_average_distance_obs_18.csv}

        \nextgroupplot[
            xticklabels=\empty,
        ]
        \plotdistance{results/segmented_long_narrow_average_distance_obs_18.csv}

        \nextgroupplot[
            xticklabels=\empty,
        ]
        \plotdistance{results/segmented_long_wide_average_distance_obs_18.csv}

        \nextgroupplot[
            hide axis,
            width=0.11\textwidth
        ]

        \nextgroupplot[
            ylabel={Distance},
            xticklabels=\empty
        ]
        \plotdistance{results/metadrive_straight_average_distance_obs_18.csv}


        \nextgroupplot[
            xlabel={Number of Samples},
            ylabel={Distance}
        ]
        \plotdistance{results/lane_shift_short_narrow_average_distance_obs_18.csv}

        \nextgroupplot[
            xlabel={Number of Samples},
        ]
        \plotdistance{results/lane_shift_short_wide_average_distance_obs_18.csv}

        \nextgroupplot[
            xlabel={Number of Samples},
        ]
        \plotdistance{results/lane_shift_long_narrow_average_distance_obs_18.csv}

        \nextgroupplot[
            xlabel={Number of Samples},
        ]
        \plotdistance{results/lane_shift_long_wide_average_distance_obs_18.csv}

        \nextgroupplot[
            hide axis,
            width=0.11\textwidth
        ]

        \nextgroupplot[
            xlabel={Number of Samples},
            ylabel={Distance},
            xtick={100,200,300,400},
            xticklabels={100,200,300,400}
        ]
        \plotdistance{results/metadrive_mixed_average_distance_obs_18.csv}

        \end{groupplot}


        \node[
            anchor=south,
            font=\small
        ] at ([yshift=7mm]
             $(plots c1r1.north)!0.5!(plots c4r1.north)$)
        {LiteRacer};

        \node[
            anchor=south,
            font=\small
        ] at ([yshift=7mm]
             $(plots c6r1.north)$)
        {MetaDrive};


        \node[
            rotate=90,
            anchor=south,
            font=\small
        ] at ([xshift=-12mm]$(plots c1r1.west)$)
        {Sinusoidal};

        \node[
            rotate=90,
            anchor=south,
            font=\small
        ] at ([xshift=-12mm]$(plots c1r2.west)$)
        {Segmented};

        \node[
            rotate=90,
            anchor=south,
            font=\small
        ] at ([xshift=-12mm]$(plots c1r3.west)$)
        {Lane Shift};


        \node[
            rotate=90,
            anchor=south,
            font=\small
        ] at ([xshift=-12mm]$(plots c6r1.west)$)
        {Curve};

        \node[
            rotate=90,
            anchor=south,
            font=\small
        ] at ([xshift=-12mm]$(plots c6r2.west)$)
        {Straight};

        \node[
            rotate=90,
            anchor=south,
            font=\small
        ] at ([xshift=-12mm]$(plots c6r3.west)$)
        {Mixed};


        \node[
            anchor=north
        ] at ([yshift=-10mm]
             $(plots c1r3.south)!0.5!(plots c6r3.south)$)
        {\pgfplotslegendfromname{commonlegend18}};

    \end{tikzpicture}

    \caption{%
        Quality of counterfactuals as a function of the number of simulation run samples for settings with $18$ obstacles, for necessary explanations (our approach), DeCaF, and random methods, measured by the average distance from the original example (lower is better).
    }
    \label{fig:counterfactual-distance-18}

\end{figure*}
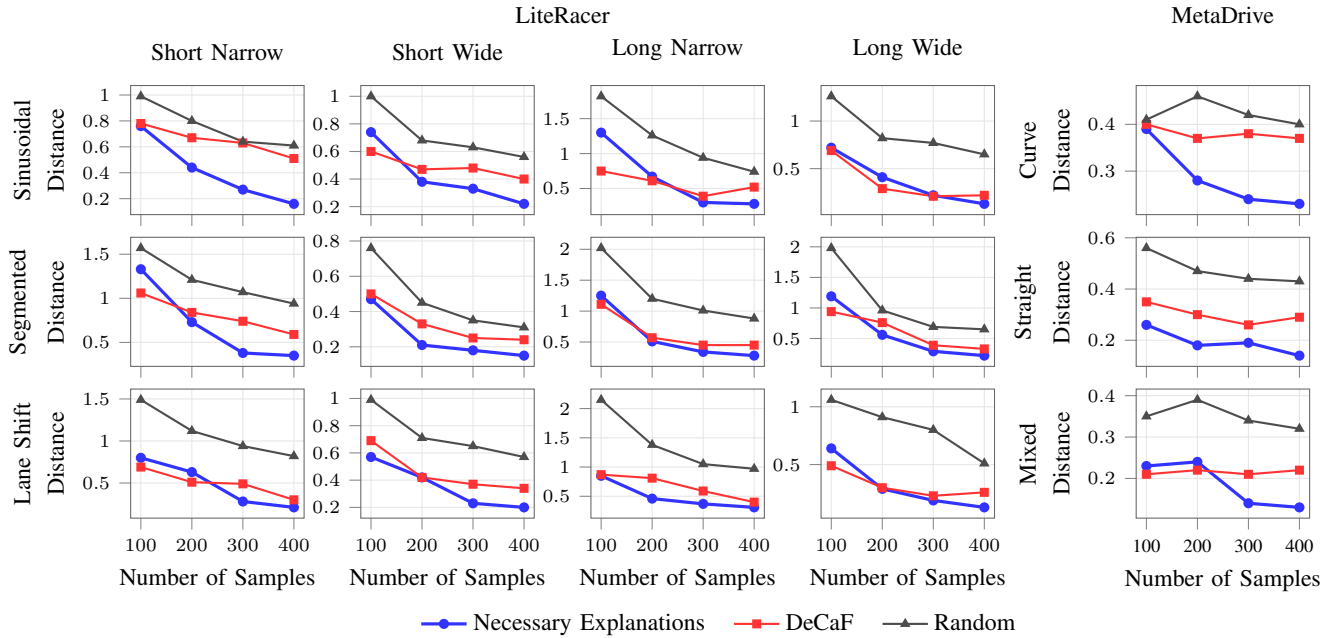

\end{appendices}


\end{document}